\documentclass[Afour,sagev,times]{sagej}
\pdfoutput=1
\usepackage{moreverb,url}

\usepackage[table,xcdraw]{xcolor}
\usepackage{multirow}
\usepackage{multicol}

\usepackage[colorlinks,bookmarksopen,bookmarksnumbered,citecolor=blue,urlcolor=blue]{hyperref}

\newcommand\BibTeX{{\rmfamily B\kern-.05em \textsc{i\kern-.025em b}\kern-.08em
T\kern-.1667em\lower.7ex\hbox{E}\kern-.125emX}}

\def\volumeyear{2019}

\graphicspath{{./images/}}

\begin{document}

\runninghead{Kyung Kim, Robert C. Leishman, and Scott L. Nykl}

\title{Virtual Testbed for Monocular Visual Navigation of Small Unmanned Aircraft Systems}

\author{Kyung Kim, Robert C. Leishman, and Scott L. Nykl}

\corrauth{Robert C. Leishman, 
Air Force Institute of Technology,
Wright-Patterson Air Force Base, OH,
US.}

\email{Robert.Leishman@afit.edu}

\begin{abstract}
Monocular visual navigation methods have seen significant advances in the last decade, recently producing several real-time solutions for autonomously navigating small unmanned aircraft systems without relying on GPS. This is critical for military operations which may involve environments where GPS signals are degraded or denied. However, testing and comparing visual navigation algorithms remains a challenge since visual data is expensive to gather. Conducting flight tests in a virtual environment is an attractive solution prior to committing to outdoor testing.


This work presents a virtual testbed for conducting simulated flight tests over real-world terrain and analyzing the real-time performance of visual navigation algorithms at 31 Hz. This tool was created to ultimately find a visual odometry algorithm appropriate for further GPS-denied navigation research on fixed-wing aircraft, even though all of the algorithms were designed for other modalities. This testbed was used to evaluate three current state-of-the-art, open-source monocular visual odometry algorithms on a fixed-wing platform: Direct Sparse Odometry, Semi-Direct Visual Odometry, and ORB-SLAM2 (with loop closures disabled).

\end{abstract}


\maketitle

\section{Introduction}
The use of small unmanned aircraft systems (SUAS) has risen in recent years across commercial and military applications. A key interest area for further development is in expanding autonomous capabilities for these systems, of
which navigation is a fundamental problem. Current navigation solutions suffer from
heavy reliance on a Global Positioning System (GPS). This dependency presents a
significant limitation, especially for critical operations that may be conducted in environments where GPS signals are degraded or actively denied. Therefore, alternative navigation solutions without GPS must be developed and monocular visual methods
are a promising and attractive approach due to the relatively low weight and energy cost of a camera along with the wealth of information available in an image. Furthermore, visual cameras are a passive sensing technology making them generally more difficult to deny than active sensing technology such as GPS.

A limitation in current monocular visual navigation research is that much of the research has focused on ground-based vehicles or quadrotor platforms flying at low altitudes and speeds in either indoor or small, constrained outdoor environments.
However, the military has a need for fixed-wing SUAS to conduct extended missions in large outdoor environments. The conditions experienced by a fixed-wing SUAS in this environment largely differ from those prior listed and the performance of visual navigation algorithms under these conditions must be explored.

Major obstacles inhibiting the ability to analyze and compare visual navigation algorithms include conducting flight tests and gathering accurate truth data. Flight tests can be a costly, time-consuming effort and some areas may be restricted or unavailable to fly over. Furthermore, gathering synchronized, accurate truth data at high frequencies is a difficult task in general and this is compounded for small flying vehicles, especially in outdoor environments.

This work addresses the issues listed above by developing a high-fidelity virtual environment for testing and analyzing visual navigation algorithms on both simulated imagery streams and real-world datasets. The virtual environment is then used to examine the performance of three current state-of-the-art visual odometry (VO) algorithms on a fixed-wing aircraft under high speeds, altitudes, and aggressive maneuvers. The algorithms tested are Direct Sparse Odometry (DSO)~\cite{Engel2018}, Semi-Direct Visual Odometry (SVO)~\cite{Forster2014svo}, and ORB-SLAM2 (with loop closures disabled)~\cite{Mur-Artal2017}. Results from simulation and real-world flight tests are also compared.

\subsection{Camera Calibration}
A necessary pre-requisite for calculating motion from images is establishing a camera model. Camera models allow mapping between a feature's 3D coordinates in the world to their corresponding 2D image coordinates. The camera model common to all three algorithms is the pinhole model, which assumes that all light rays incident to an image pass through single point~\cite{Hartley2004}. Cameras must be calibrated such that the image information conforms to the pinhole model assumption. The calibration finds the intrinsic parameters, including the focal lengths and image center coordinates and the lens distortion parameters. A popular method for calibrating cameras, and the method used in this work, involves taking pictures of a flat checkerboard with known dimensions from multiple positions and orientations.

\subsection{Visual Navigation Categories}
Visual navigation methods can largely be divided into two classes: VO and visual simultaneous localization and mapping (VSLAM). VO incrementally computes the relative motion of a vehicle or camera from a sequence of images and tracks a more localized environment~\cite{Scaramuzza2011b, Scaramuzza2011}. VSLAM falls under the larger umbrella of SLAM which seeks to maintain a global map of the environment and a globally consistent trajectory, which are used to update and correct one other\cite{Durrant-whyte2006, bailey06simultaneous}. 

Additionally, VO and VSLAM algorithms can both be categorized based on their image comparison approach, which use either feature-based or direct methods to compare two or more images and calculate the transformation describing the motion between the images. 
Feature-based approaches~\cite{Hassaballah2016} identify key points of interest throughout images and extract feature vector representations that are invariant to changes in conditions, such as lighting, rotation, translation, and scale. Robust features are necessary to be able to calculate and find the same representations when viewed from different perspectives. Features are matched across images and the transformations between the images are calculated by optimizing the geometric error of the feature positions. 
Direct methods~\cite{Irani1999} avoid the costly steps of extracting and matching features, and instead directly calculate the transformation that optimizes the photo-metric error, or the difference in pixel intensities, between images. 

Finally, VO and VSLAM algorithms differ by the pixel density of images used to calculate motion. The pixel densities are sparse, semi-dense, and dense methods. Sparse methods use a few points throughout images, semi-dense methods use regions of pixels, and dense methods use every pixel in the images.

\subsection{Related Work}
Parallel Tracking and Mapping (PTAM)~\cite{Klein2007} laid a foundation architecturally for modern VO and VSLAM algorithms. PTAM introduced the idea of splitting tracking and mapping into separate threads and using a keyframe-based approach to allow real-time bundle adjustments for higher overall performance and accuracy.

Dense Tracking and Mapping (DTAM)~\cite{Newcombe2011} is a fully dense and direct algorithm which utilizes a Graphics Processing Unit (GPU) to achieve real-time performance. Using every pixel in an image, DTAM was shown to be significantly more robust under occlusions, camera blur, and defocus than other sparse, feature-based methods. However, most VO and VSLAM solutions for SUAS and other mobile robotics currently focus on implementing solely CPU-based algorithms to minimize the required processing power.

Large-Scale Direct SLAM (LSD-SLAM)~\cite{Engel2014lsd} is a semi-dense VSLAM algorithm using direct methods for pose estimation and a loop closure method for correcting trajectories when the camera appears to be viewing a scene that was previously observed. 

DSO~\cite{Engel2018} is a fully direct and sparse VO algorithm. DSO performs optimizations over a sliding window of keyframes and optionally incorporates a photometric camera calibration in addition to a geometric calibration. A photometric calibration models the camera's exposure response and pixel attenuation (vignetting)~\cite{Engel2016}. This calibration can be beneficial for direct methods as they operate directly on pixel brightness intensities but is unnecessary for feature-based methods since features are usually invariant to changes in lighting conditions.

ORB-SLAM is one of the leading feature-based VSLAM algorithms~\cite{Mur-Artal2015}. ORB-SLAM uses three threads: one each for tracking, mapping, and loop closing. One of the greatest strengths of ORB-SLAM is the efficiency of using the same features for tracking, re-localization, and loop detection. The authors improved upon the original work\cite{Mur-Artal2015} and created ORB-SLAM2~\cite{Mur-Artal2017}. ORB-SLAM2 adds in a full bundle adjustment step over all keyframes and map points in a fourth thread. It also extends the algorithm to accommodate monocular, stereo, and/or RGB-D cameras.

SVO~\cite{Forster2014svo} is a hybrid VO algorithm that extracts features only on keyframes to initialize new map points, but uses a direct method to calculate motion for every frame. This allows SVO to achieve high frame rates. SVO has been used as a baseline algorithm in multiple works. REMODE~\cite{Pizzoli2014} is a monocular dense reconstruction algorithm which uses a GPU to achieve real-time performance. REMODE uses SVO to estimate the camera's pose. Follow on methods~\cite{Faessler2015} fused SVO outputs with IMU measurements to autonomously control a quadrotor and execute a trajectory through an onboard processor while providing a real-time dense 3D map of the traversed area using REMODE on an offboard computer. SVO 2.0~\cite{Forster2017} extends the original SVO algorithm to large field-of-view cameras, multi-camera systems, IMU incorporation, and the additional use of edges for feature alignment. However, the original SVO algorithm was used in this work since SVO 2.0 is currently only available as a precompiled binary and the source code is not publicly available.

A relative navigation architecture~\cite{Wheeler2017} consists of a relative front-end running a multi-sate constraint Kalman filter for state estimation and a global back-end maintaining a pose graph for optimizations and loop closures. The closest work to that described herein is a relative navigation method for fixed wing UAVs~\cite{Ellingson2018}. They describe a simulation environment that evaluates Wheeler's relative navigation architecture. The simulated aircraft was flown over a cityscape image at 11 m/s and an altitude of 50 m. However, the algorithms was not yet fast enough to run under real-time constraints and the frame rate of the virtual aircraft camera was limited to three frames-per-second (fps).

Recently, new approaches to visual navigation have been explored involving deep learning and artificial neural networks. DeepVO~\cite{Wang2017DeepVO:Networks} uses deep recurrent and convolutional neural networks to calculate pose estimates from image sequences. This method makes camera calibrations and geometric optimizations unnecessary and is able to recover absolute scale without external input, as all this is learned while training the neural net.  However, these methods can require large amounts of data to train and make functional.  Publicly available training data for flights of fixed-wing UAVs is limited.  
Several public VO and VSLAM datasets with synchronized truth data are available to test algorithm performances.  The KITTI dataset~\cite{Geiger2013} contains data collected from a station wagon driving through traffic at 10 fps. The TUM RGB-D dataset~\cite{Sturm2012} consists of color and depth images from a Microsoft Kinect sensor at 30 fps in an indoor office environment from either a handheld configuration or on a wheeled robot.  The TUM monoVO dataset~\cite{Engel2016} demonstrates 50 sequences of monocular data exploring both indoor and outdoor environments from a handheld configuration. However, the ground truth was generated through LSD-SLAM and is not completely accurate.  Others~\cite{Burri2016a} presented the European Robotics Challenge Micro Aerial Vehicle (MAV) dataset with a hexrotor flying in an indoor industrial environment. Ground truth data was recorded with either a laser tracking system at 20 Hz with millimeter accuracy or a motion capture system at 100 Hz.  Finally, the Zurich Urban MAV dataset~\cite{Majdik2016} is based on a quadrotor flying outdoors over the streets of Zurich, Switzerland at altitudes of 5-15 m over a 2 km trajectory. However, ground truth data was generated using appearance-based topological localization and VSLAM algorithms. Each dataset described above contains inherent limitations and, most importantly, are largely inapplicable to fixed-wing SUAS flying in large outdoor environments. 

Simulation environments for visual odometry are becoming more common.  The increased interest is primarily due to the large data requirements for training deep-learning approaches for visual processing, e.g. \cite{gaidon2016virual}.  Researchers have found that even large, real-world datasets (e.g. KITTI \cite{Geiger2013}) are insufficient to satisfy the insatiable needs of training a deep neural network.   There are a few approaches in the literature that are similar to the proposed approach, however none are focused on the needs and particularities of fixed-wing UAVs.   

The approach outlined in \cite{sayre2018visual} presents a methodology for creating impressive photo-realistic image generation that is then processed by a VO pipeline.  Estimates are then used within the control loop of the UAV.  The approach is geared toward rotorcraft UAVs in small environments.  

Microsoft has released an open-source simulation environment, known as AirSim \cite{shah2018airsim}, built using the Unity game engine. The environment supports driving vehicles and multirotor UAVs and can stream out imagery and point clouds from virtual cameras/scanners mounted on platforms. The platform currently does not support fixed wing UAVs. The Unity engine has been used in other efforts for similar purposes \cite{zhang2016benefit} 

Gazebo \cite{koenig2004design} is the go-to simulation environment for robotics research, supported by the ROS middleware. Several packages have been built to support multirotor unmanned vehicles, such as RotorS \cite{furrer2016rotors}, but the environment is lacking in support for fixed-wing vehicles.    


\begin{figure*}[h]
\centering
\includegraphics[trim=0.1in 2.8in 0.1in 0.1in, clip, width=\textwidth]{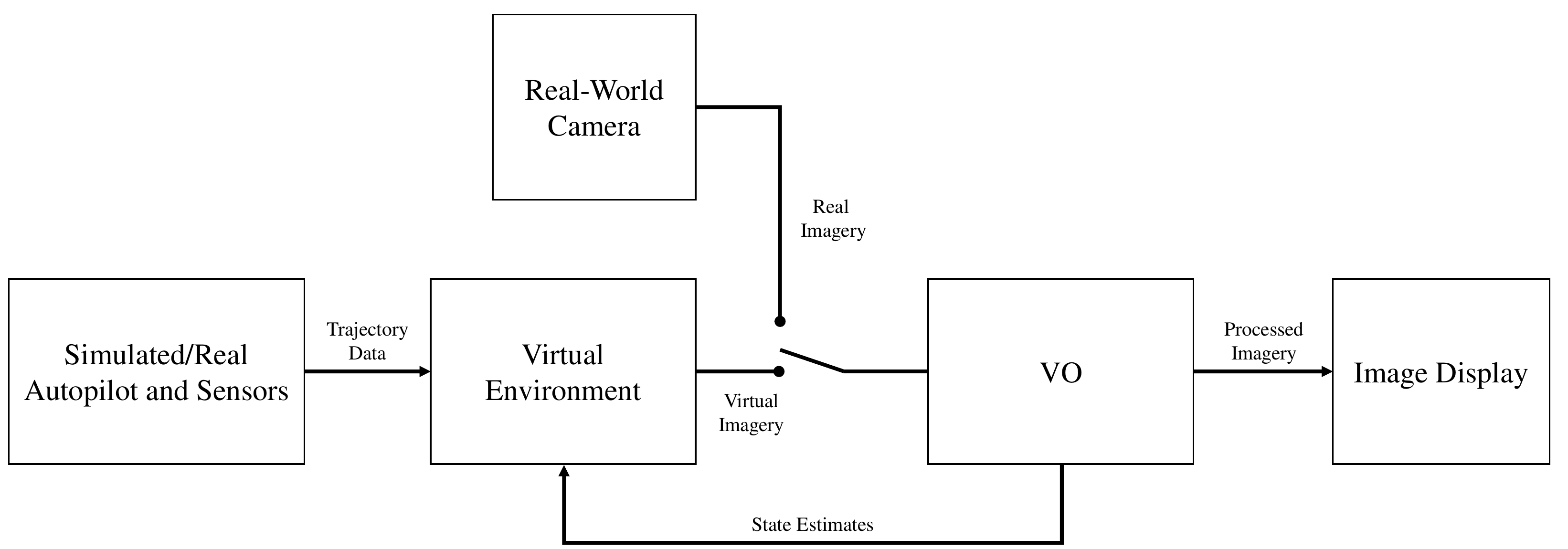}
\caption{\label{fig:autonomy_sim_overview}Navigation Testbed System. The navigation testbed is a flexible and modular virtual environment that allows both simulated and real-world flight profiles to run in real-time over GPS-terrain-mapped satellite imagery. Estimated trajectories from VO algorithms are projected in real-time and can be compared to one another and to truth. The simulator is also able to record and play back trajectories.}
\end{figure*}

This work presents a novel virtual testbed for conducting simulated flight tests over real-world terrain and for playing back previously recorded flight data.  The real-time performance of visual navigation algorithms can then be analyzed and compared simultaneously to the performance of other algorithms within the same virtual testbed.  The virtual environment was built using the AftrBurner engine, a cross-platform visualization engine \cite{Nykl2008}. The engine has been used successfully for testing stereo computer vision techniques in simulation for automated aerial refueling applications~\cite{Johnson2017} and for testing structure from motion~\cite{nykl2018_boomOcclusion, roeber2018_SfM,parsonsAAR_AIAA_IS}.

\section{Simulation System}
This work makes two contributions by using a novel collection of existing open source tools combined with a couple new capabilities developed as part of this effort.  The first contribution is an ability to generate synthetic, physically correct trajectories and imagery for fixed-wing SUAS vehicles.  The second is a pluggable framework optimized for testing and evaluating visual odometry algorithms, which features a 3D visualization capability to compare algorithm performance.  This section will document the simulation system. First an overview of the framework will be presented and then each major component will be discussed in some detail. We outline below the sources for the different existing open source components and the novel additions that combine to form the entire system.  

\subsection{System Overview}
 
The navigation testbed is a flexible and modular virtual environment that allows both simulated and real-world flight profiles to run in real-time over GPS-terrain-mapped satellite imagery. Estimated trajectories from VO algorithms are projected in real-time and can be compared to one another and to truth. The simulator is also able to record and play back trajectories. 

\begin{figure*}
\centering
\includegraphics[width=\textwidth]{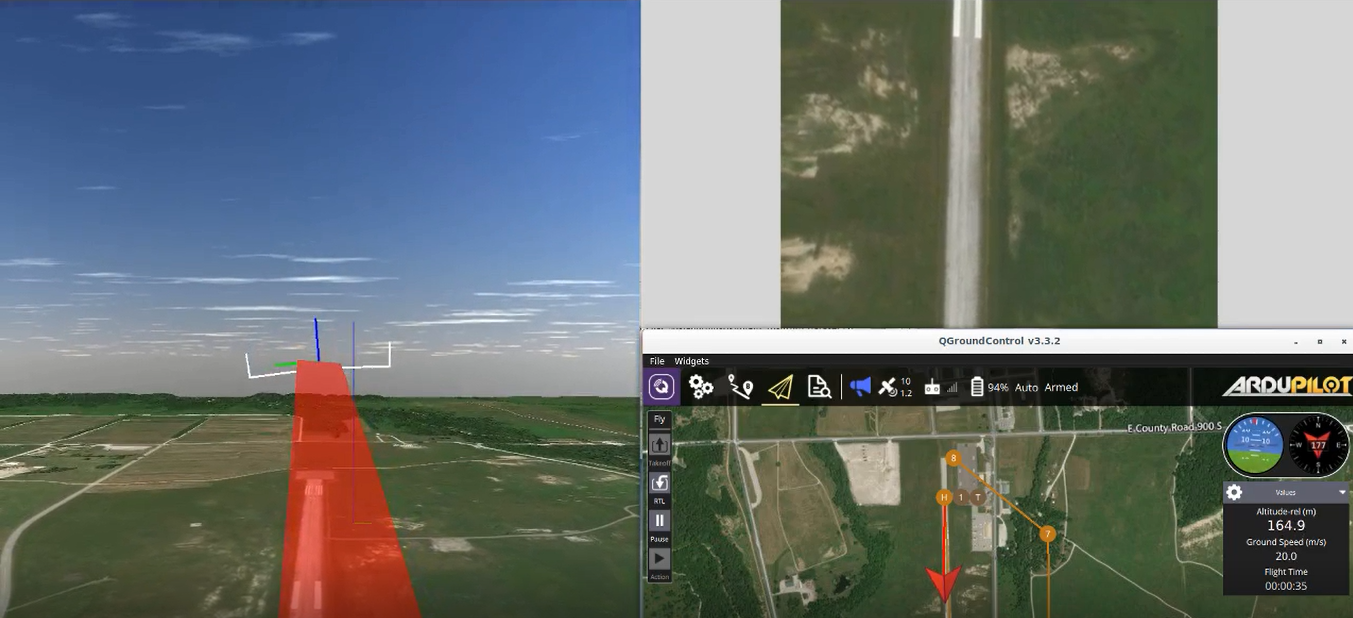}
\caption{\label{fig:sim_view}Simulator view when using completely simulated information. The left window shows the virtual environment, with georegistered imagery draped on DEM tiles. The top right window shows the live image feed from the virtual camera attached to the bottom of the aircraft. The bottom right window shows the ground control system used to plan flights and control the aircraft.}
\end{figure*}

Figure~\ref{fig:autonomy_sim_overview} illustrates a high-level flow diagram of the system and
%
Figure~\ref{fig:sim_view} shows the components of the overall navigation testbed system. The terrain is composed of satellite imagery augmented with elevation data. A SUAS's trajectory data is generated through a simulated or real-world autopilot. Corresponding sensor output can be projected into the virtual environment. This projection enables one to visualize what the SUAS is actually sensing and place the data in the context of the SUAS's current geographical position and orientation. 

The virtual environment produces imagery from a virtual camera attached to the SUAS flying a desired trajectory. This imagery is then used as input for a VO algorithm. Alternatively, the VO algorithms can be used to process imagery streams from a real-world camera. The VO results are then fed back into the virtual environment to allow live visualization and analysis of the VO's 6 Degree-of-freedom performance. Each of the VO algorithms also provide the processed state of the input imagery showing information such as which pixels or features were selected and used. This can be displayed through a separate node.

\subsection{Autopilot Interface}
The ArduPilot\footnote{\url{https://ardupilot.org/dev/index.html}}  open-source autopilot software is used to control the aircraft in both the virtual environment and the real world. ArduPilot uses an EKF to fuse sensor input and provide low-level control for various autonomous vehicles including rovers, submarines, copters, and planes. Raw and fused sensor data can be read from the ArduPilot. 

ArduPilot provides a Software In The Loop (SITL) program which allows ArduPilot software to be run on a computer and simulate virtual autopilots for vehicles. This is used to generate simulated trajectories for a fixed-wing aircraft exhibiting realistic flight dynamics. 

A ground control station (GCS) program is required to give commands and communicate with the ArduPilot. The GCS used for controlling the simulated aircraft is QGroundControl since it is available on Linux, allowing simultaneous execution of all modules on the same laptop computer. The GCS used for controlling the real-world aircraft is Mission Planner since that is the standard GCS used in the ANT Center.

The ArduPilot uses the MAVLink protocol to transmit and receive data. This data can be streamed or logged.  The simulation system can connect directly to a real or SITL autopilot or load logged autopilot data.

\subsection{Virtual Environment}
The virtual environment box in Figure~\ref{fig:sim_view} is actually composed of several components.  The primary compoent is the AftrBurner Engine, which then loads several other modules as plugins.  These compoenents are detailed below.  The novel software contributions of this work were created as plugins to the AftrBurner engine. 

\subsection{AftrBurner Engine}
\begin{figure}
\centering
\includegraphics[width=0.5\textwidth]{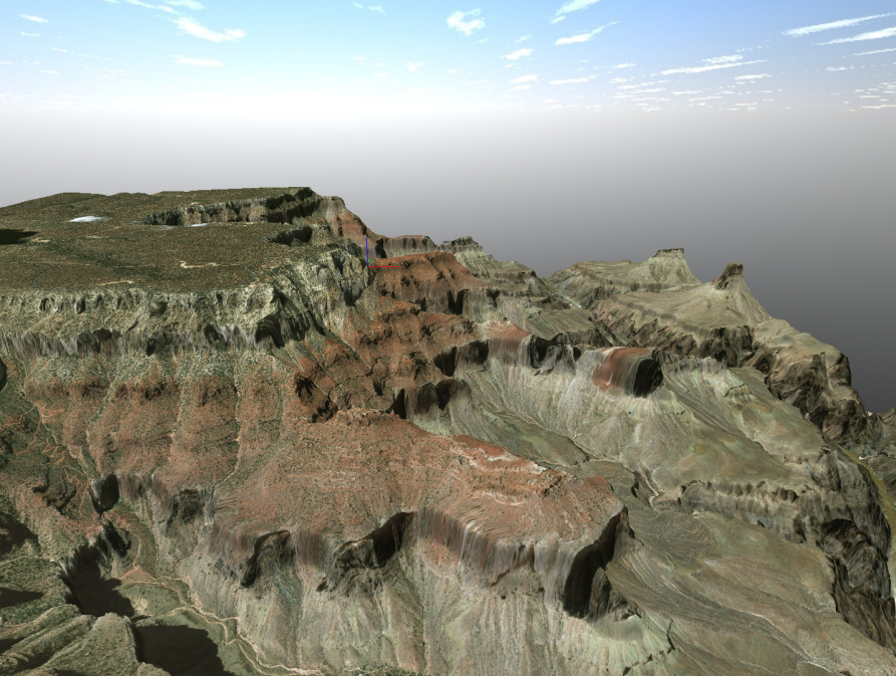}
\caption{\label{fig:virtual_grand_canyon}Virtual Grand Canyon terrain as an example of the capability of the AftrBurner engine. Note that the distortion caused by the large changes in elevation would have consequences for the VO algorithms and, consequently, such terrain was not employed for this study.}
\end{figure}

All virtual environment components are built using the AftrBurner engine, a cross-platform visualization engine written in C++ and the successor to the STEAMiE educational game engine \cite{Nykl2008}. The visualization engine contains submodules for creating and reading virtual camera data as well as generating scaled real-world terrain models using United States Geological Survey (USGS) elevation data and satellite imagery. The AftrBurner engine also contains formulations for modeling the World Geodetic System (WGS) 84 model to accurately visualize GPS coordinates. Figure \ref{fig:virtual_grand_canyon} shows an example terrain model of the Grand Canyon created through the AftrBurner engine mapped against GPS coordinates and USGS elevation data. Limiting factors include the available resolution of terrain and satellite imagery and the image distortion incurred from mapping a 2D overhead image onto a 3D model. This distortion is more pronounced in extreme elevation changes as shown on the cliffs and walls of the virtual Grand Canyon.  We did not test over areas of extreme elevation changes in this work.  The smearing and distortion present in Figure \ref{fig:virtual_grand_canyon} would cause accuracy issues with the VO algorithms tested.  

The AftrBurner engine is the core of the Virtual Environment and most of the capabilities used in this work are off the shelf from the available plugins.  However, the following are new additions, created for this work. 

\subsection{Camera Calibration}
\begin{figure}
\centering
\includegraphics[width=0.5\textwidth]{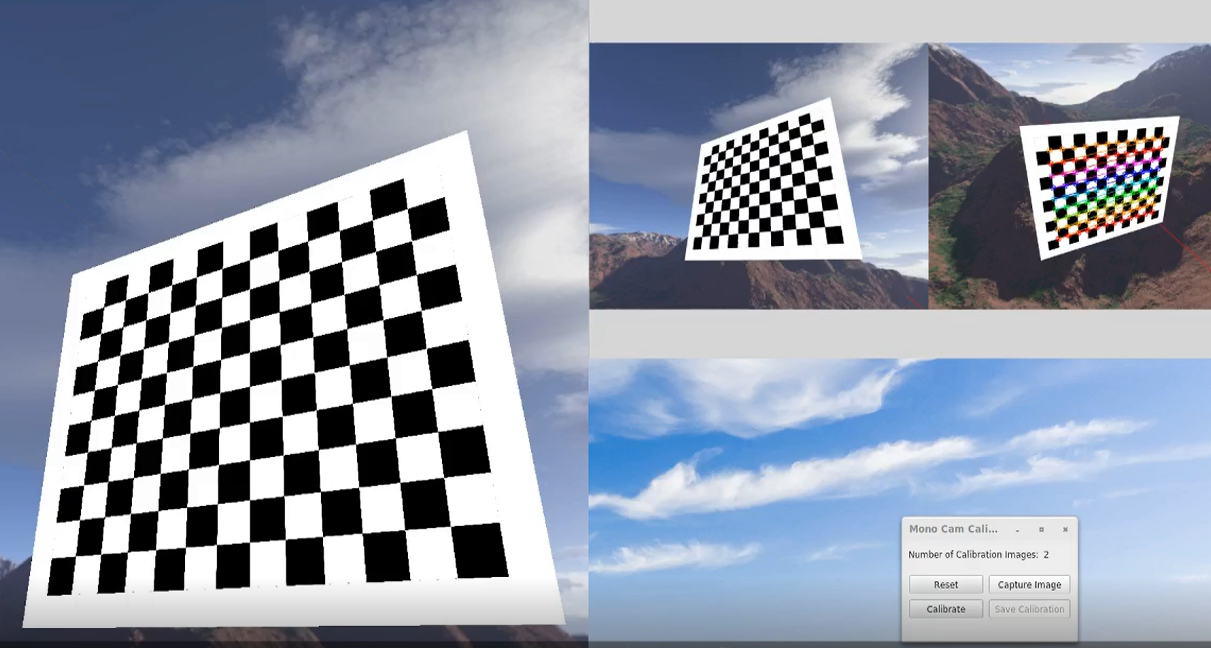}
\caption{\label{fig:cam_calibration}The camera calibration module. The left window shows a camera in the virtual environment viewing a checkerboard. The right windows show the calibration module.}
\end{figure}

A camera calibration module was constructed to extract the intrinsic parameters of a monocular camera using the pinhole camera model within the AftrBurner Engine.  The module accepts imagery streams from a camera viewing a checkerboard as input and works interchangeably with virtual camera imagery as well as real-world camera imagery streams. Figure \ref{fig:cam_calibration} shows an example of using the camera calibration module with a virtual camera. The left window shows the perspective of the camera in the virtual environment. The right windows show the camera calibration module which presents both a live stream of the input imagery received and the last image captured for calibration. The calibration results are then used in the VO algorithms.

\subsection{Trajectory Viewer}
\begin{figure*}
\centering
\includegraphics[trim = 2.0in 10.0in 14.0in 6.0in, clip, width=\textwidth]{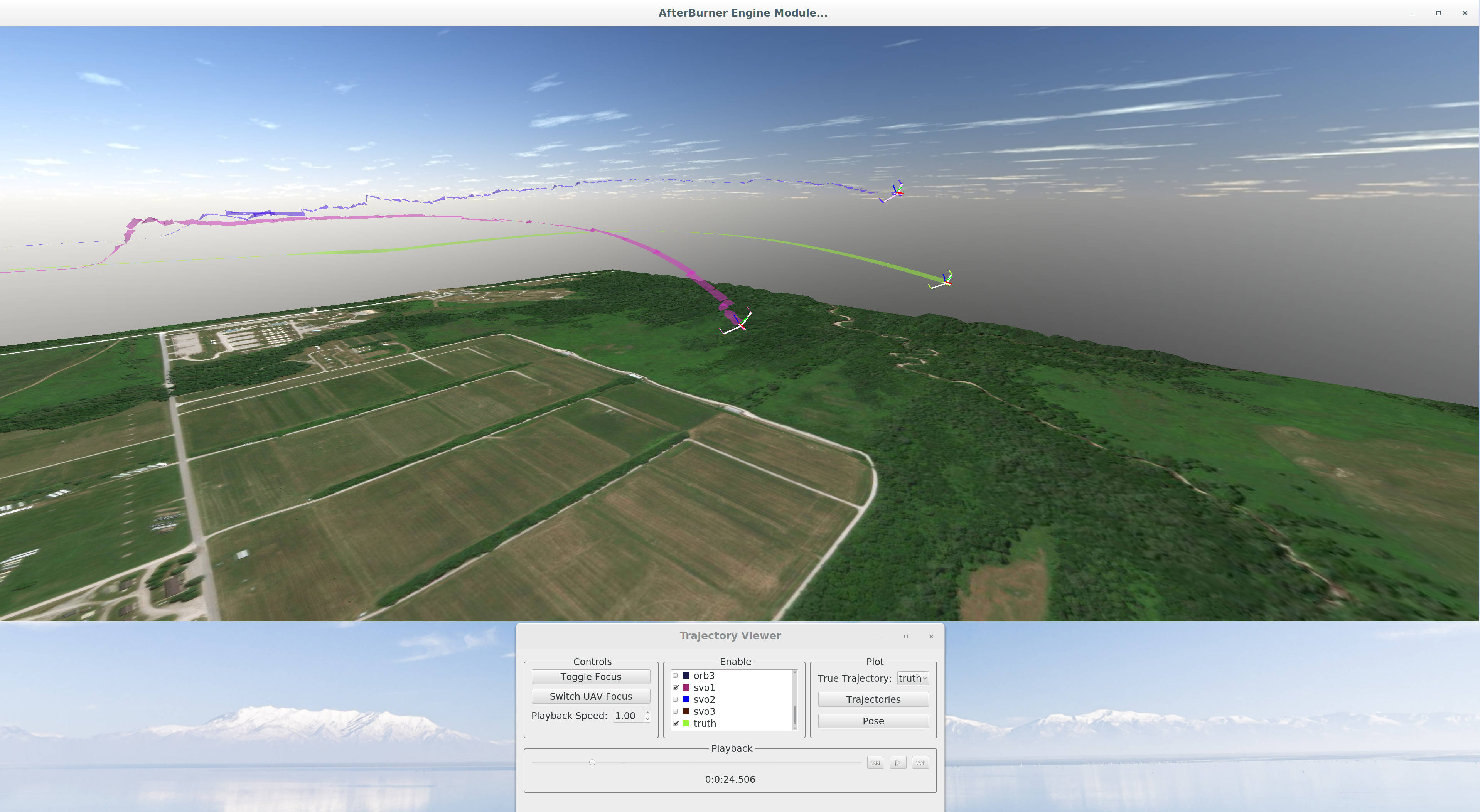}
\caption{\label{fig:trajectory_viewer}Here is a view of the simulated world with the trajectory viewer enabled. The ribbons outline the trajectories of the different VO estimates.}
\end{figure*}

A trajectory viewer module is used to view multiple trajectories simultaneously.  This plugin was developed as part of this work and is original. This module is primarily for playback of a single truth trajectory and multiple estimated trajectories from VO algorithms applied to that truth trajectory. Trajectory playback is synchronized across all trajectories, can be skipped to any point in time and sped up or slowed down. 3D plots of the trajectory and 2D plots for x, y, z, roll, pitch, and yaw data can also be generated. This allows detailed analysis and visualization of algorithm performances in all six degrees of freedom and to scale. Figure \ref{fig:trajectory_viewer} shows an example of the trajectory viewer displaying multiple trajectories. 

\begin{figure}
\centering
\includegraphics[width=0.5\textwidth]{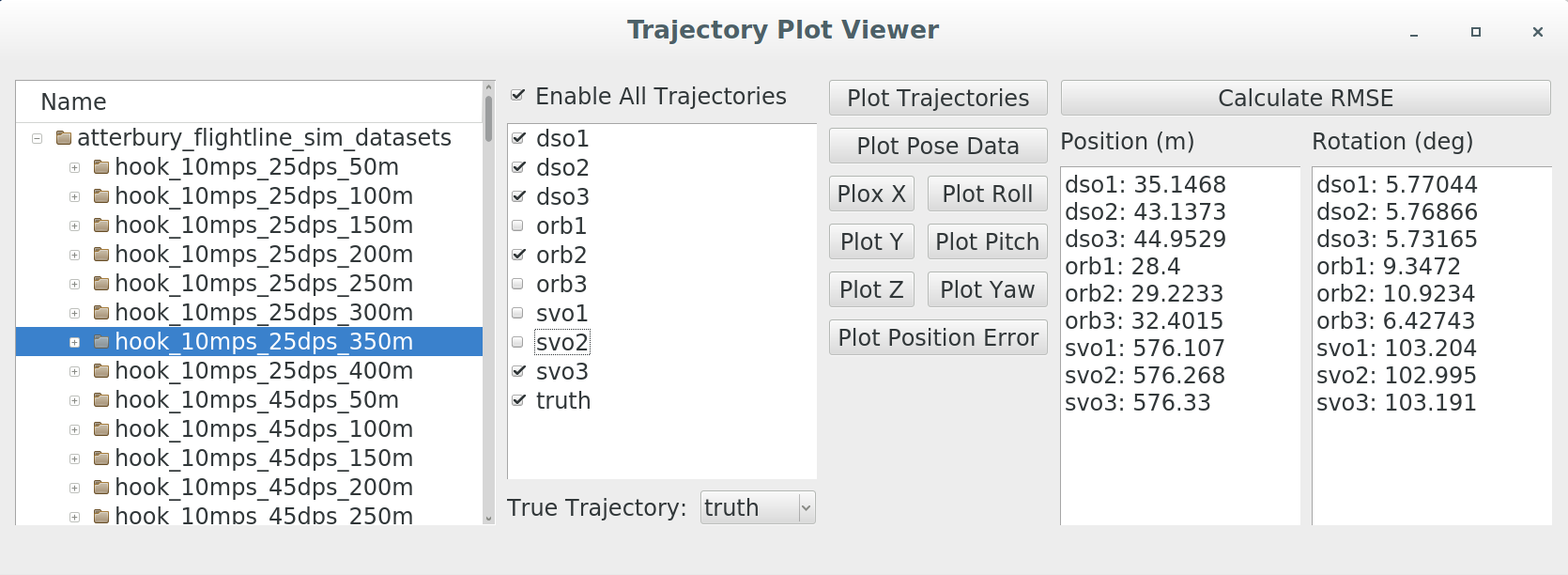}
\caption{\label{fig:trajectory_viewer_control}Here is a view of the trajectory viewer control panel. }
\end{figure}

A current limitation of the trajectory viewer is that it can only load one dataset during each execution. Therefore, an additional trajectory plot viewer module is provided to allow analysis and plotting of multiple datasets in one process. It also provides more extensive plotting capabilities to be able to inspect individual plots in more detail as well as calculate and print out the trajectory position and rotation errors. Figure \ref{fig:trajectory_viewer_control} shows the controls for the trajectory plot viewer. The left-most window pane allows the selection of directories containing truth and estimated trajectories for a given flight. Selecting a directory auto-populates the next window pane with the trajectories to allow selective plotting. The truth trajectory file must also be identified since it is used to synchronize the estimated trajectory files to ensure that pose plots are generated against the same flight times. The next window panes contain controls for plotting and calculating the position and rotation errors for each of the trajectory files.

\subsection{Visual Odometry Algorithms}
The Navigation Testbed is configurable such that any VO algorithm that can take imagery via ROS or LCM should be pluggable within the system.  Three different open-source, state-of-the-art algorithms are tested: DSO \cite{Engel2018}, SVO \cite{Forster2014svo}, and ORB-SLAM2 \cite{Mur-Artal2017}. SVO is used instead of SVO2 since the source code for SVO2 is not currently open-source. ORB-SLAM2 is allowed for comparison against the other VO algorithms by disabling the loop closure thread so that it essentially functions as a feature-based VO algorithm.


Each of the VO algorithms are tested using their default parameter settings. Additional tests are conducted by adjusting individual parameters specific to each algorithm. In DSO, setting\_minGradHistAdd controls the image gradient threshold for candidate point selection in an image. Tests are run using the default value of seven and a lowered value of five to allow the selection of points with a lower gradient since many of the scenes viewed over Camp Atterbury's flightline have little texture \cite{Engel2018}. The FAST score thresholds of ORB-SLAM2 and SVO are also tested at lower values for the same reason. ORBextractor.minThFAST in ORB-SLAM2 and triangMinCornerScore in SVO are tested at their default values of seven and a lowered value of two \cite{Forster2014svo} \cite{Mur-Artal2015}. 

SVO controls the creation of new keyframes through the kfSelectMinDist parameter which is the percentage of the average scene depth. A new frame is designated as a keyframe if it is farther than this distance from the previous keyframes. This is tested at its default value of 12\% and a lowered value of 5\% to create keyframes more frequently and increase the robustness of the algorithm under fast movements.  Additional details for each algorithm are found below.  

\subsubsection{DSO}
DSO was released by Engel et al. in 2016 \cite{Engel2018}. This algorithm not only utilizes geometric camera calibration to model the camera's intrinsic parameters, but it also incorporates an optional photometric camera calibration to model the camera's exposure response and pixel attenuation (vignetting). Photometric camera calibration is not necessary in feature-based methods since features are selected to be invariant to changes in lighting conditions but is helpful in direct methods since they depend solely on pixel intensities. In DSO, new frames are tracked relative to the latest keyframe. If the frame is not designated as a new keyframe, it is discarded. If a new keyframe is created, the photometric error is optimized over a sliding window of the latest seven keyframes. DSO tracks a fixed, sparse set of pixels across all keyframes. Candidate pixels in an image are selected for tracking by dividing the image into blocks of size $d \times d$ and calculating a region-adaptive gradient threshold for each block based on the median gradient of the pixels in the block. The pixel with the highest gradient greater than the threshold is chosen. To be able to include pixels in regions with small gradients, the process is repeated again over larger block sizes of $2d$ and $4d$ with weaker gradient thresholds.  DSO is currently open-source.

\subsubsection{ORB-SLAM2}
ORB-SLAM is one of the leading feature-based VSLAM algorithms and was created by Mur-Artal et al. in 2015 \cite{Mur-Artal2015}. ORB-SLAM involves three threads for tracking, mapping, and loop closing. ORB-SLAM maintains an undirected weighted graph of keyframes that are linked to each other based on the number of shared map points in the keyframe images. This covisibility graph is used for loop closures and pose graph optimizations. However, in order to increase efficiency, a subset of the covisibility graph is also maintained as a spanning tree and is called the essential graph. The essential graph contains all keyframe nodes but only connects keyframes sharing the most map points. 

ORB-SLAM extracts ORB features from an image and uses them for tracking, relocalization, and loop detection which allows for high efficiency. In the tracking thread, an initial pose estimation is obtained for the current frame by either using a constant velocity motion model from the last frame if tracking was successful in the last frame or, if tracking was lost, by relocalizing from the bag-of-words recognition database. The pose estimation is optimized by projecting the local map of keyframes containing covisible map points into the current frame. The keyframe with the most covisible map points is designated as the reference keyframe. If the current frame's image is sufficiently different from that of the reference keyframe, it is inserted into the covisibility graph as a new keyframe in the mapping thread. Map points that have been determined to be non-trackable or erroneously triangulated are culled. Otherwise, new map points are triangulated from matching ORB features in connected keyframes. Local bundle adjustment is run to optimize the current keyframe along with the connected keyframes in the covisibility graph and all map points belonging to those keyframes. Redundant keyframes are then culled. 

The loop closing thread utilizes the bag-of-words approach to query the recognition database, find similar images, and insert new edges into the covisibility graph for loop closures. The loop closure error is then propagated throughout the graph by optimizing the essential graph and transforming all map points according to its keyframe's correction. A new version of the algorithm, ORB-SLAM2, was released in 2017  \cite{Mur-Artal2017}. ORB-SLAM2 adds in a full bundle adjustment step over all keyframes and map points in a separate fourth thread. It also extends the algorithm to accommodate monocular, stereo, and RGB-D cameras. ORB-SLAM and ORB-SLAM2 are both open-source.

\subsubsection{SVO}
SVO is a hybrid VO algorithm by Forster et al. that was first released in 2014 \cite{Forster2014svo}. SVO consists of a motion estimation or tracking thread and a mapping thread. In the motion estimation thread, an initial estimate of the camera pose is found through sparse model-based image alignment in which a direct method is used to minimize the photometric error with reprojected feature patches from the previous image. The reprojected features in the new image are then aligned with respect to the rest of the map by optimizing the 2D pixel locations to minimize the photometric error with the reference feature patch in the keyframe. By adjusting each individual reprojected feature locations in the new frame, the feature alignment step violates epipolar constraints. In order to correct this, SVO performs a bundle adjustment through the pose and structure refinement step, optimizing the camera pose again but this time by minimizing the photometric error with respect to optimized feature patches. If the mapping thread receives a keyframe, it splits the image into fixed-size cells and extracts the FAST corners with the highest Shi-Tomasi score in each cell. Depth filters are initialized for new features with high uncertainty. New regular frames are used to update these depth filters using a Bayesian method and once the variance is sufficiently low, the corresponding 3D point is inserted into the map to be used for motion estimation. By only extracting feature points during keyframes and using direct methods to calculate the camera pose for every frame, SVO is able to achieve high processing speeds. The authors also presented SVO 2.0 in 2017 \cite{Forster2017}. This extends the original SVO algorithm to large FOV cameras, multi-camera systems, IMU incorporation, and the additional use of edges for feature alignment. Although the original SVO algorithm is open-source, SVO 2.0 is only available as a pre-compiled binary.

\subsection{Message Protocols}
A core issue for robotic systems is inter-process communication (IPC) since most systems require the development and use of multiple subsystems running as separate processes or nodes. Several open-source middleware have been developed to address this problem and one of the most popular is the Robot Operating System (ROS) \cite{Quigley2009}. IPC in ROS is accomplished through a message-passing system where language-agnostic message types can be built and used to generate the appropriate language-specific data structures and files to be imported by nodes for communication. Messages are passed from publisher to subscriber nodes that are on the same topic. Key design features in ROS are that all communications between nodes are managed by a centralized node and that the principal protocol used is the Transmission Control Protocol (TCP). This makes ROS less than ideal for building distributed systems with potentially unreliable connections between nodes. However, ROS features an extensive mature library beyond IPC with a large open-source community, provides a centralized parameter server and launch system for easy management and distribution of data among nodes, and maintains a build system augmenting CMake for support in importing ROS packages and libraries. 

Another open-source middleware is the Lightweight Communications and Marshalling (LCM) library \cite{Huang10LCM}. This was designed by students at MIT to offer the bare essentials of just IPC. LCM operates using a similar system as ROS, passing messages from publisher nodes to subscriber nodes that are on the same channel. Communication between nodes is accomplished through the User Datagram Protocol (UDP) in a decentralized network, making LCM more suitable for distributed systems. However, LCM lacks the presence of ROS's large open-source community and libraries. LCM also lacks a centralized parameter server, requiring alternative methods to distribute parameters to disparate nodes.

The lines depicted in Figure~\ref{fig:autonomy_sim_overview} represent message passing through one of these two transport protocols. ROS is used for its launch system and parameter server to be able to correctly initialize and launch multiple nodes. It is also used to take advantage of the MAVROS library, an extensive and mature library for communicating with the ArduPilot autopilot and simulator. LCM is used for compatible communication with systems being developed at the AFIT Autonomy and Navigation Technology (ANT) Center. ROS Lunar and LCM version 1.3.1 are used in this work.

\subsection{Implementation Details} \label{sec:implementation_details}
All modules were built and tested on a laptop computer with an Intel Core i7-7500U processor and a Nvidia Quadro M2200 GPU. The operating system used was Ubuntu 16.04. All modules were programmed in C++. Inter-Process Communication (IPC) between the individual nodes were accomplished mainly through the Robotic Operating System (ROS)~\cite{Quigley2009}. However, IPC with the real-world autopilot, sensors, and camera were accomplished through the Lightweight Communications and Marshalling (LCM) library~\cite{Huang10LCM}.

Simulated trajectories are generated by interfacing the virtual environment with the MAVROS library and Ardupilot autopilot software. Ardupilot provides a software-in-the-loop (SITL) module which simulates an autopilot. SITL can be programmed with missions and controlled through a standard ground control station (GCS) to conduct simulated flight tests with realistic flight dynamics. Figure \ref{fig:sim_view} shows an example in which a GCS is used to plan a flight and control the aircraft. The left window shows the virtual environment, the top right window shows the live image feed from the virtual camera attached to the bottom of the aircraft, and the bottom right window shows QGroundControl, the GCS.

\section{Experiments}
Flight tests were conducted both in simulation and in the real world over Camp Atterbury, Indiana with a fixed-wing aircraft and a camera facing straight downwards relative to the aircraft's body frame. The objectives of the flight tests are two-fold: demonstrate the utility of the virtual world flights in evaluating visual navigation approaches and to identify the best state-of-the-art VO algorithm for use on a fixed-wing UAV. These objective were evaluated by analyzing the real-time performance of each algorithm under high speeds and altitudes while performing aggressive maneuvers.

\subsection{Assumptions}
A consideration that has to be taken into account is the inability of monocular visual navigation algorithms to be able to determine absolute scale without external input. All three algorithms compute trajectory positions to a relative scale. To account for this, all flight tests included an initialization leg in which the aircraft flies straight and level such that the camera points straight down towards the Earth. The VO algorithms are initialized during this leg and the altitude at which the algorithm is initialized is used as the absolute scale for the entire trajectory. Furthermore, the loop closure capability in ORB-SLAM2 was disabled in order to make a fair comparison against the other strict VO algorithms.

Both SVO and DSO were able to process images and produce state estimates at the frame rate of the input imagery stream, 31 fps. However, ORB-SLAM2 was limited to producing state estimates at a rate of 20 fps. 

A small search over several different settings and parameters for each VO algorithm was conducted~\cite{Kim2019Monocular}.  The default parameters for each of the algorithms offered the best performance in both the simulation and real-world flight tests and all reported results found below were obtained from using the default settings.

\subsection{Simulation Setup}
\begin{figure}[t]
\centering
\includegraphics[width=0.5\textwidth]{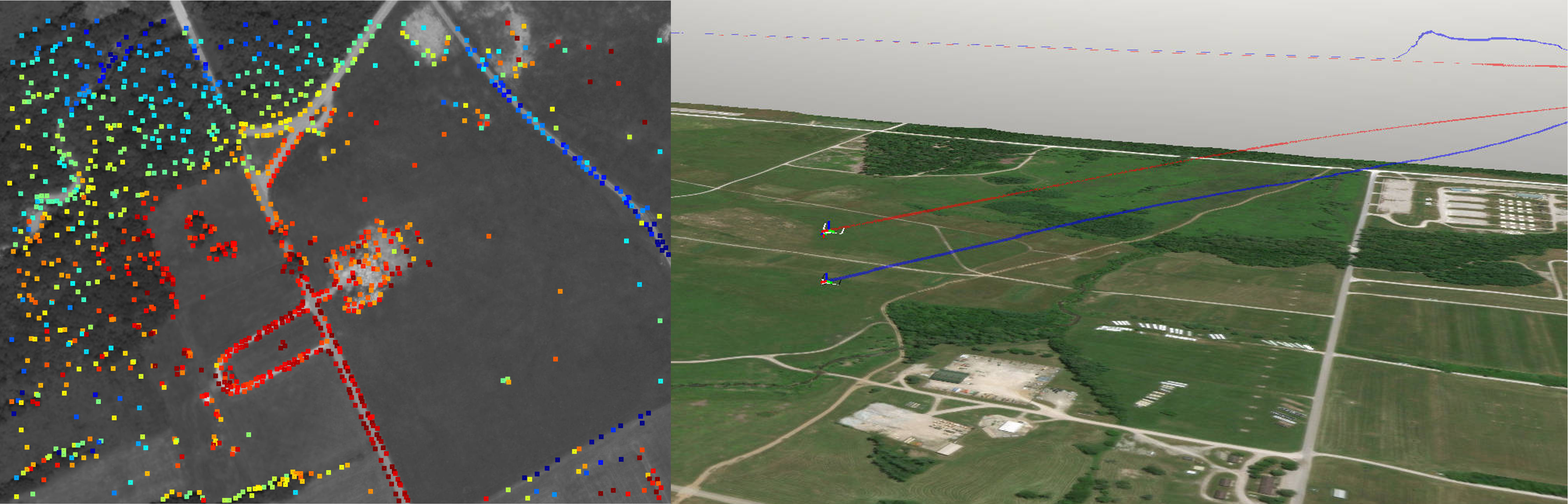}
\caption{\label{fig:dso_simulated}Simulated flight with DSO.  The left image is an output of DSO, showing the points of interest for the navigation. The right image shows the virutal world and the comparison between truth and estimates. The red SUAS ribbon shows the truth trajectory and the blue shows the estimate from DSO.}
\end{figure}

\begin{figure}[t]
\centering
\includegraphics[trim= 0.0in 4.0in 0.0in 4.5in, clip, width=0.5\textwidth]{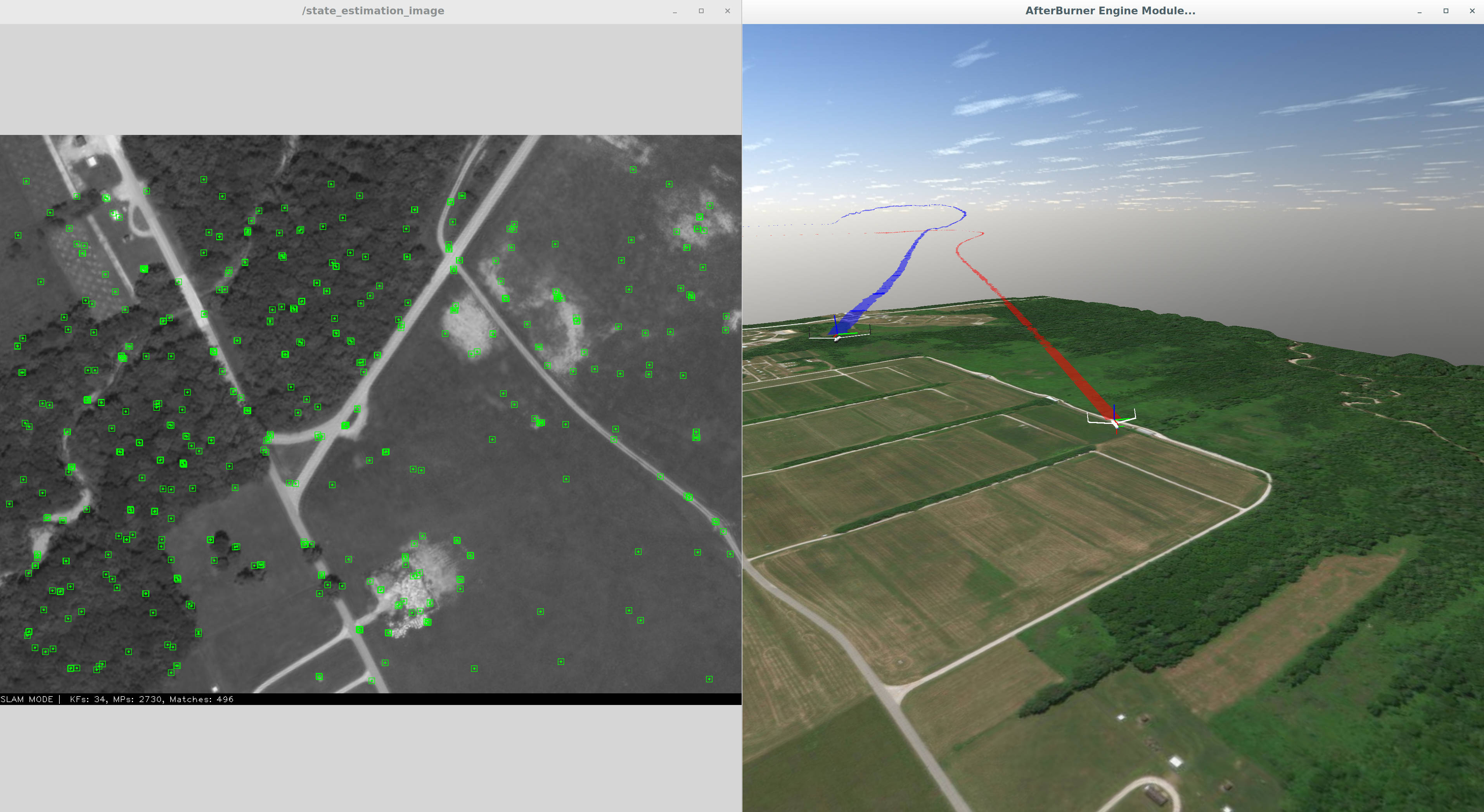}
\caption{\label{fig:orb_simulated}Simulated flight with ORB-SLAM2. The left image is an output of DSO, showing the points of interest for the navigation. The right image shows the virutal world and the comparison between truth and estimates. The red SUAS ribbon shows the truth trajectory and the blue shows the estimate from ORB-SLAM2.}
\end{figure}

\begin{figure}[t]
\centering
\includegraphics[trim= 0.0in 4.0in 0.0in 4.5in, clip, width=0.5\textwidth]{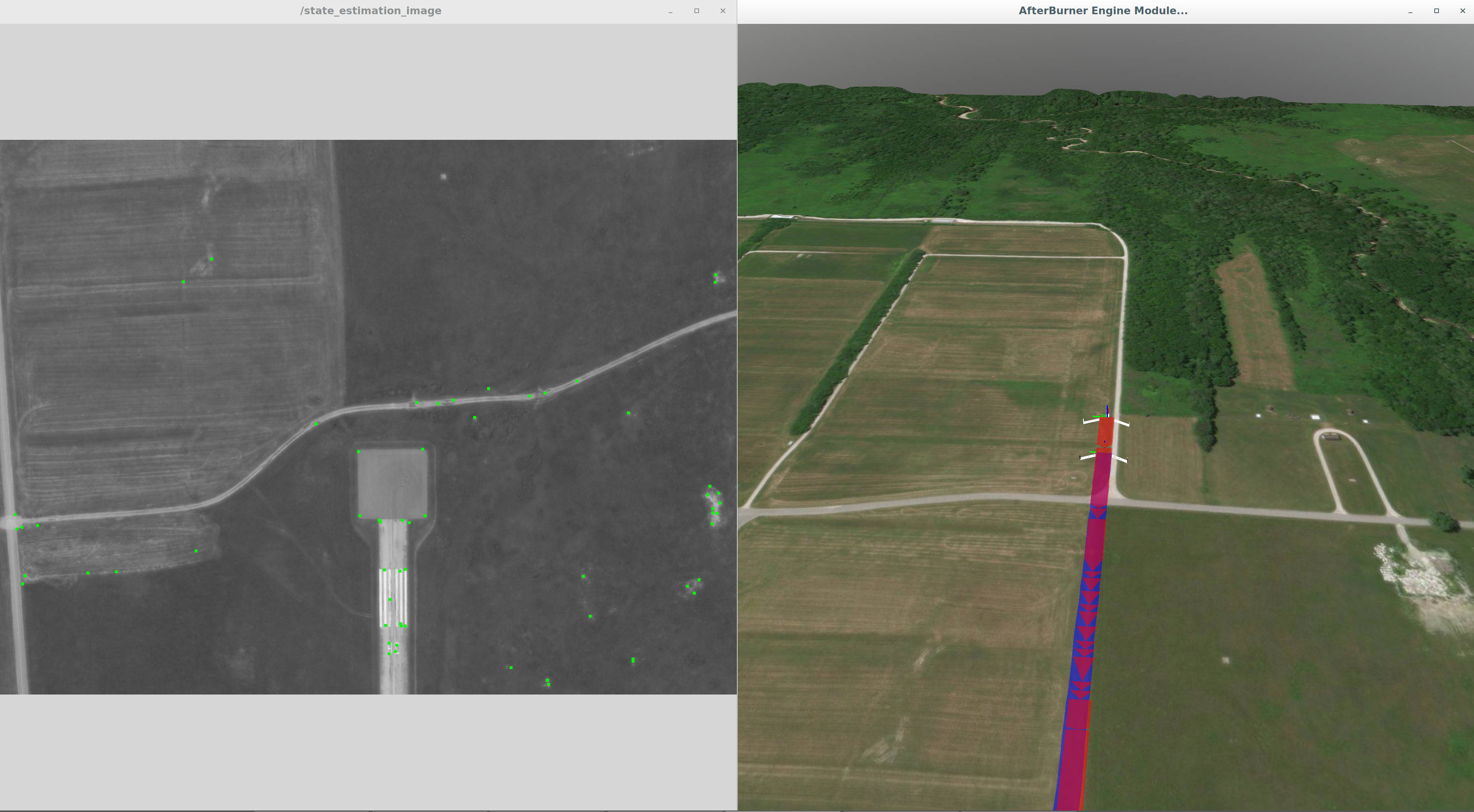}
\caption{\label{fig:svo_simulated}Simulated flight with SVO. The left image is an output of DSO, showing the points of interest for the navigation. The right image shows the virutal world and the comparison between truth and estimates. The red SUAS ribbon shows the truth trajectory and the blue shows the estimate from SVO.}
\end{figure}

Figures \ref{fig:dso_simulated}, \ref{fig:orb_simulated} and \ref{fig:svo_simulated} show examples of running each of the algorithms on simulated flight trajectories and reprojecting the pose estimates back into the virtual environment. The left windows show the processed images from the VO algorithms with the tracked pixels or features in the current image. The right windows show the virtual environment with the truth trajectory represented by the SUAS with the red trailing ribbon and the estimated trajectory by the SUAS with the blue trailing ribbon.

The virtual camera settings used are shown in Table \ref{tab:virtual_cam_settings}.  The virtual environment was able to continuously supply a framerate of 31 Hz of simulated imagery during all the simulation experiments discussed below. 

\begin{table}[h]
\small\sf\centering
\caption{Virtual Camera Settings.\label{tab:virtual_cam_settings}}
\begin{tabular}{ll}
\toprule
Option & Setting\\
\midrule
Resolution & 1280 $\times$ 960\\
Aspect Ratio (width/height) & 1.333\\
Field of View (deg) & 84.872\\
Frame rate (fps) & 31\\
\bottomrule
\end{tabular}
\end{table}

\begin{figure}
\centering
\includegraphics[width=0.5\textwidth]{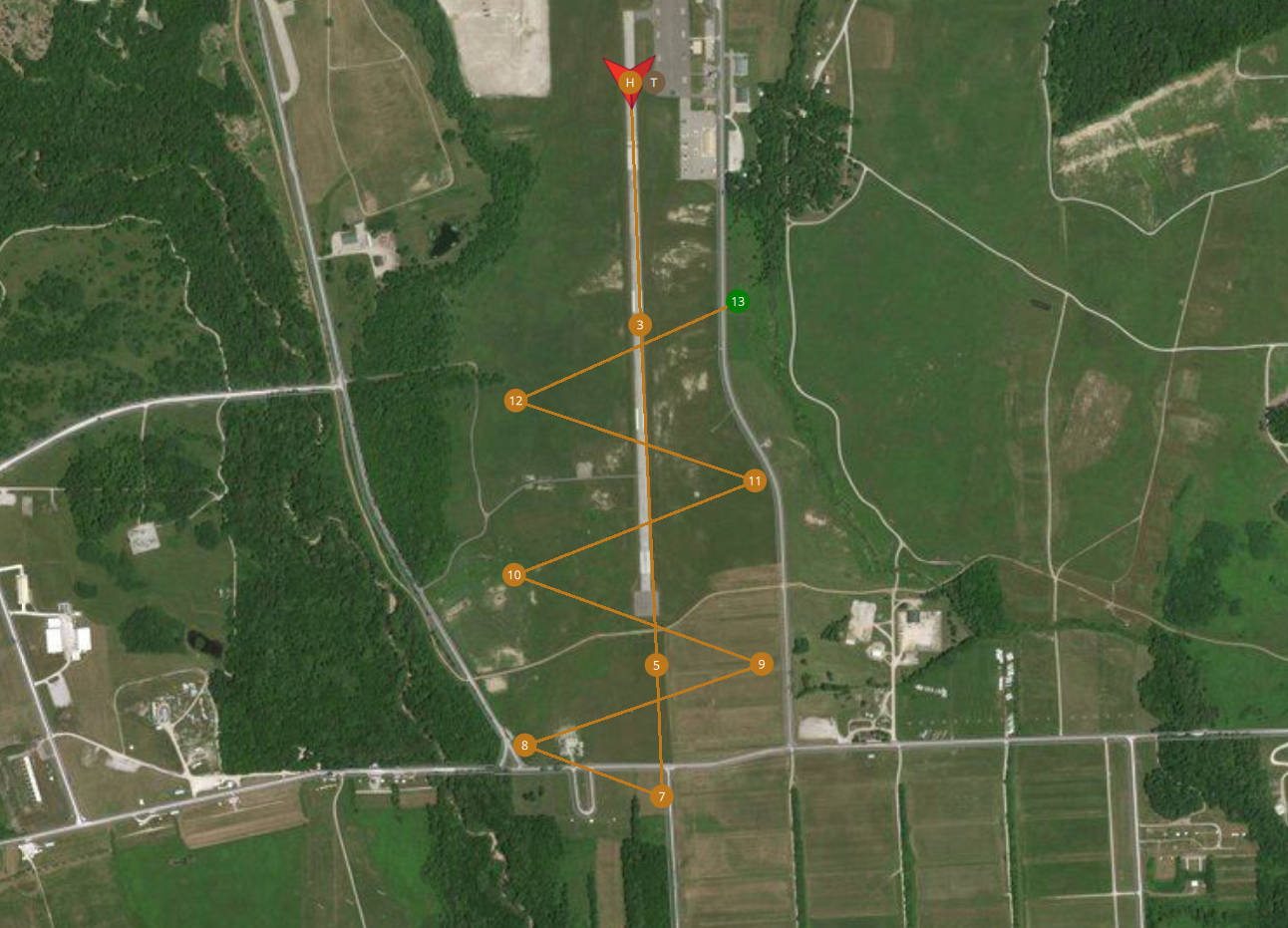}
\caption{\label{fig:oscillate_sim_trajectory}One sample of a simulation trajectory use during this study.  The results shown are from this trajectory.  Additional trajectories examined are found in }
\end{figure}

The algorithms were tested on a 3.3 km trajectory that is shown in Figure \ref{fig:oscillate_sim_trajectory}, where the SUAS starts at the north end of the runway, flies south for the initialization leg, and then returns north in an oscillating pattern. The experimental parameters that were varied are shown in Table \ref{tab:sim_experimental_params}.  In this simulation, we did not add wind and the trajectory is relatively flat, so airspeed is equal to ground speed.  The roll rate setting limits the maximum roll rate allowed by the SITL autopilot.   

\begin{table}[h]
\small\sf\centering
\caption{Simulation Experimental Parameters.\label{tab:sim_experimental_params}}
\begin{tabular}{ll}
\toprule
Experimental Parameters & Settings\\
\midrule
Airspeed (m/s) & 10, 25\\
Maximum roll rate (deg/s) & 25, 45, 65\\
Altitude & 50-400 m in 50 m intervals\\
\bottomrule
\end{tabular}
\end{table}

\subsection{Real-World Flight Test}
The VO algorithms were also tested on data gathered from real-world flight tests on a 1/3-scale Carbon Cub aircraft weighing 50 lbs with a wingspan of 14 ft. A Pixhawk 2 autopilot was used for low-level control of the aircraft and all payload data was stored in LCM log files for post-flight playback and analysis. The LCM log files were recorded in-flight on an onboard Intel NUC7 computer with an i7 processor. Grayscale imagery was collected from a Prosilica GT1290 camera whose properties are listed in Table \ref{tab:real_cam_properties}. Truth data was provided by a Piksi navigation board with GPS data at 10 Hz and IMU data at 100 Hz. All data was timestamped using a TM2000A Precision Time Protocol (PTP) server. Processing of the flight data was conducted on the laptop computer.

\begin{table}[h]
\small\sf\centering
\caption{Real-World Camera Properties.\label{tab:real_cam_properties}}
\begin{tabular}{ll}
\toprule
Property & Setting\\
\midrule
Resolution & 1280 $\times$ 960\\
Aspect Ratio (width/height) & 1.333\\
Lens Field of View (deg) & 84.872\\
Lens Sensor Format (in) & 1/2\\
Camera Sensor Format (in) & 1/3\\
Frame rate (fps) & 33\\
\bottomrule
\end{tabular}
\end{table}

\begin{figure}[h]
\centering
\includegraphics[width=0.5\textwidth]{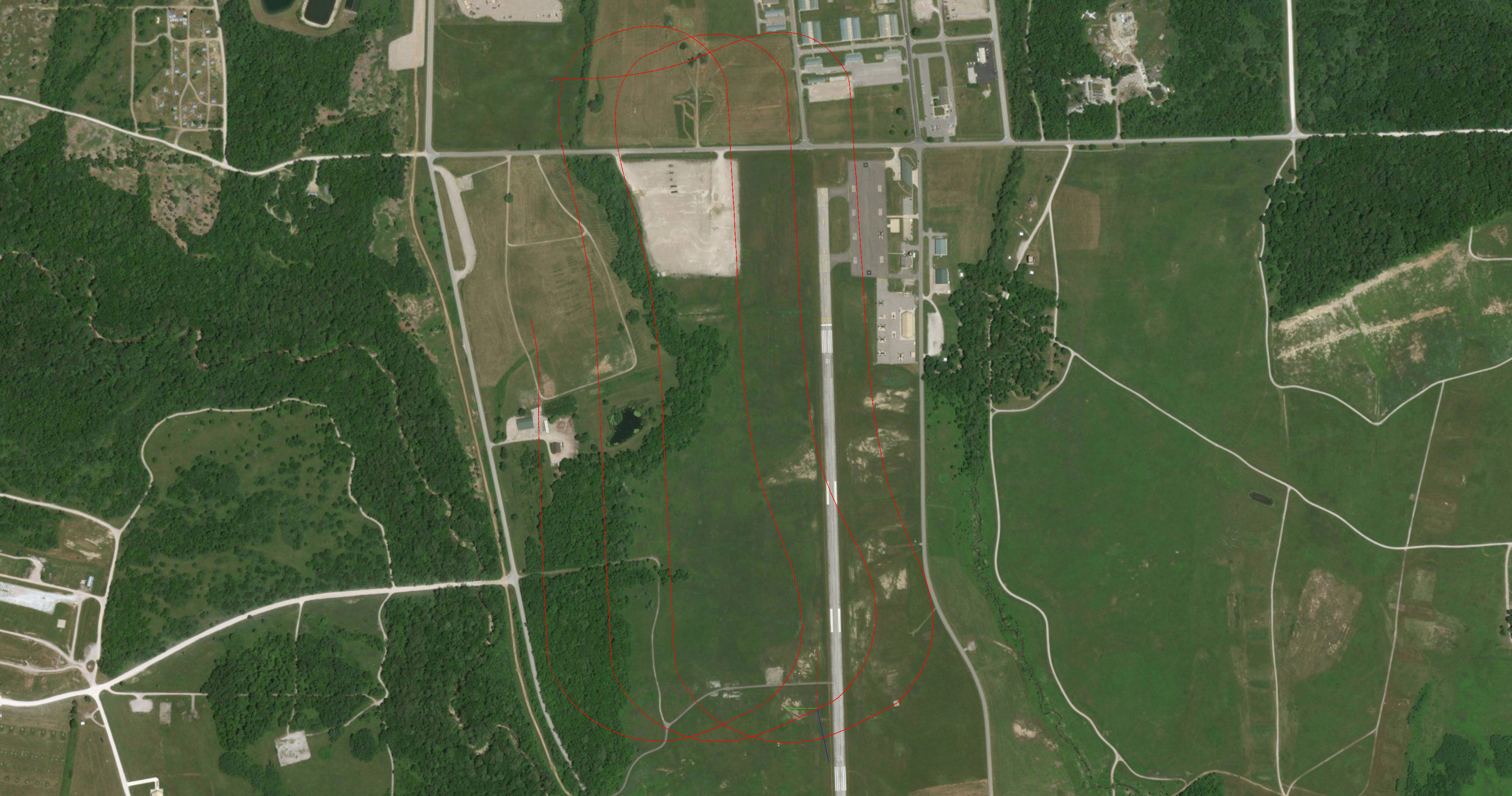}
\caption{\label{fig:real_world_truth_trajectory}Path flown for the real-world results.}
\end{figure}

The algorithms were tested on a 9.6 km trajectory shown in Figure \ref{fig:real_world_truth_trajectory} where the aircraft starts at the western end of the airfield and flies three counter-clockwise loops while moving east. The trajectory was flown at 250 m in altitude with airspeeds varying between 20-30 m/s for a flight time of 284 seconds.   

The weather the day of the flight testing provided additional challenges.  Much of the imagery from the few successful flights is close to saturation due the exposure time set on the camera to achieve its maximum frame rate of 33 fps. This leads to a saturated image on bright days which, when combined with the low texture environment of the airfield, provides a difficult challenge for the VO algorithms to be able to detect and track features or pixels. We did not attempt to recreate the saturated images from the flight test with the simulation environment, as one of the main objectives of the study was to evaluate visual odometry performance.  However, the capabilities of the simulation would have allowed for that possibility. Future work will examine simulating different sensor failures to evaluate for robustness of algorithms when faced with challenging sensor information.   

\section{Experimental Results}
This section discusses the simulation results first and then the real-world flight results. Algorithm performance is reported through the 3D root-mean-square error (RMSE) of the position in meters and the rotation in degrees from the ground truth throughout the entire trajectory.   Overall, ORB-SLAM2 proved to be the most flexible and robust algorithm throughout a variety of conditions in both simulation and real-world flight tests.  

\subsection{Simulation}
The results for the simulation flights with a slower airspeed of 10 m/s are shown in Table \ref{tab:default_10mps}. If an algorithm failed to initialize and track throughout the entire trajectory, the position and rotation errors are marked with an X. Additionally, the entries for the algorithm with the lowest position and rotation errors for a trajectory are highlighted. Most of the algorithms encountered difficulties initializing at lower altitudes due to the limited resolution of the satellite imagery and the limited field-of-view. 

SVO had the most difficulty and was unable to initialize below 300 m in altitude. SVO also proved to be the most fragile algorithm and experienced great difficulty maintaining localization, especially during rotations. This is due to the method of keyframe selection. SVO creates new keyframes strictly based on the Euclidean distance between the newest frame and the previous keyframes, failing to take rotations into account. For this reason, SVO was never able to track past the first turn. 

DSO and ORB-SLAM2 were able to successfully track the entire trajectory for the majority of the test cases. ORB-SLAM2 provided the most accurate position estimates throughout all trajectories in which it was able to initialize. Although ORB-SLAM2 and DSO produced comparable rotation estimates with ORB-SLAM2 being slightly more accurate at lower roll rates, DSO provided a more consistent and accurate estimate at the highest roll rate of 65 deg/s.

\begin{table}[h]
\centering
\caption{\label{tab:default_10mps}Simulation VO Results at 10 m/s. The position and rotation errors for trajectories in which a VO algorithm failed to initialize are marked with an X. The algorithm with the lowest position or rotation error for a trajectory is highlighted.}
\resizebox{0.5\textwidth}{!}{
\begin{tabular}{|c|c|c|c|c|c|c|c|}
\hline
                                                                                       &                                                                                   & \multicolumn{3}{c|}{\textbf{Position RMSE (m)}}                                 & \multicolumn{3}{c|}{\textbf{Rotation RMSE (deg)}}                               \\ \cline{3-8} 
\multirow{-2}{*}{\textbf{\begin{tabular}[c]{@{}c@{}}Roll Rate\\ (deg/s)\end{tabular}}} & \multirow{-2}{*}{\textbf{\begin{tabular}[c]{@{}c@{}}Altitude\\ (m)\end{tabular}}} & \textbf{DSO}                    & \textbf{ORB}                   & \textbf{SVO} & \textbf{DSO}                   & \textbf{ORB}                  & \textbf{SVO} \\ \hline
                                                                                       & \textbf{50}                                                                       & X                               & X                              & X            & X                              & X                             & X            \\ \cline{2-8} 
                                                                                       & \textbf{100}                                                                      & \cellcolor[HTML]{FFFE65}170.997 & X                              & X            & \cellcolor[HTML]{FFFE65}4.075  & X                             & X            \\ \cline{2-8} 
                                                                                       & \textbf{150}                                                                      & X                               & \cellcolor[HTML]{FCFF2F}60.898 & X            & X                              & \cellcolor[HTML]{FCFF2F}4.914 & X            \\ \cline{2-8} 
                                                                                       & \textbf{200}                                                                      & 113.955                         & \cellcolor[HTML]{FCFF2F}28.420 & X            & \cellcolor[HTML]{FCFF2F}5.541  & 64.059                        & X            \\ \cline{2-8} 
                                                                                       & \textbf{250}                                                                      & 110.229                         & \cellcolor[HTML]{FCFF2F}20.434 & X            & 6.504                          & \cellcolor[HTML]{FCFF2F}5.470 & X            \\ \cline{2-8} 
                                                                                       & \textbf{300}                                                                      & 127.009                         & \cellcolor[HTML]{FCFF2F}20.137 & 403.935      & 7.965                          & \cellcolor[HTML]{FCFF2F}7.113 & 188.684      \\ \cline{2-8} 
                                                                                       & \textbf{350}                                                                      & 126.387                         & \cellcolor[HTML]{FCFF2F}18.376 & 545.184      & 8.274                          & \cellcolor[HTML]{FCFF2F}5.717 & 187.95       \\ \cline{2-8} 
\multirow{-8}{*}{\textbf{25}}                                                          & \textbf{400}                                                                      & 132.861                         & \cellcolor[HTML]{FCFF2F}15.331 & 539.435      & 12.257                         & \cellcolor[HTML]{FCFF2F}6.784 & 182.351      \\ \hline
                                                                                       & \textbf{50}                                                                       & X                               & X                              & X            & X                              & X                             & X            \\ \cline{2-8} 
                                                                                       & \textbf{100}                                                                      & \cellcolor[HTML]{FCFF2F}152.962 & X                              & X            & \cellcolor[HTML]{FCFF2F}5.101  & X                             & X            \\ \cline{2-8} 
                                                                                       & \textbf{150}                                                                      & 104.605                         & \cellcolor[HTML]{FCFF2F}93.130 & X            & \cellcolor[HTML]{FCFF2F}26.227 & 28.169                        & X            \\ \cline{2-8} 
                                                                                       & \textbf{200}                                                                      & 136.505                         & \cellcolor[HTML]{FCFF2F}28.474 & X            & 10.358                         & \cellcolor[HTML]{FCFF2F}5.355 & X            \\ \cline{2-8} 
                                                                                       & \textbf{250}                                                                      & 106.315                         & \cellcolor[HTML]{FCFF2F}14.406 & X            & 8.429                          & \cellcolor[HTML]{FCFF2F}6.415 & X            \\ \cline{2-8} 
                                                                                       & \textbf{300}                                                                      & 146.665                         & \cellcolor[HTML]{FCFF2F}15.811 & 428.557      & 11.835                         & \cellcolor[HTML]{FCFF2F}5.923 & 184.938      \\ \cline{2-8} 
                                                                                       & \textbf{350}                                                                      & 119.18                          & \cellcolor[HTML]{FCFF2F}17.349 & 520.012      & \cellcolor[HTML]{FCFF2F}7.558  & 22.937                        & 198.616      \\ \cline{2-8} 
\multirow{-8}{*}{\textbf{45}}                                                          & \textbf{400}                                                                      & 176.371                         & \cellcolor[HTML]{FCFF2F}15.749 & 535.812      & 17.827                         & \cellcolor[HTML]{FCFF2F}6.922 & 183.166      \\ \hline
                                                                                       & \textbf{50}                                                                       & X                               & X                              & X            & X                              & X                             & X            \\ \cline{2-8} 
                                                                                       & \textbf{100}                                                                      & \cellcolor[HTML]{FCFF2F}182.635 & X                              & X            & \cellcolor[HTML]{FCFF2F}83.158 & X                             & X            \\ \cline{2-8} 
                                                                                       & \textbf{150}                                                                      & 124.558                         & \cellcolor[HTML]{FCFF2F}77.100 & X            & \cellcolor[HTML]{FCFF2F}27.253 & 84.419                        & X            \\ \cline{2-8} 
                                                                                       & \textbf{200}                                                                      & 131.615                         & \cellcolor[HTML]{FCFF2F}15.688 & X            & \cellcolor[HTML]{FCFF2F}6.216  & 6.455                         & X            \\ \cline{2-8} 
                                                                                       & \textbf{250}                                                                      & 83.809                          & \cellcolor[HTML]{FCFF2F}18.574 & X            & \cellcolor[HTML]{FCFF2F}25.237 & 70.235                        & X            \\ \cline{2-8} 
                                                                                       & \textbf{300}                                                                      & 138.955                         & \cellcolor[HTML]{FCFF2F}15.458 & 527.49       & 11.463                         & \cellcolor[HTML]{FCFF2F}6.142 & 186.732      \\ \cline{2-8} 
                                                                                       & \textbf{350}                                                                      & 133.632                         & \cellcolor[HTML]{FCFF2F}19.362 & 514.676      & \cellcolor[HTML]{FCFF2F}18.412 & 72.723                        & 185.18       \\ \cline{2-8} 
\multirow{-8}{*}{\textbf{65}}                                                          & \textbf{400}                                                                      & 154.969                         & \cellcolor[HTML]{FCFF2F}16.639 & 516.447      & \cellcolor[HTML]{FCFF2F}23.975 & 26.571                        & 192.534      \\ \hline
\end{tabular}}
\end{table}

\begin{figure}[t]
\centering
\includegraphics[width=0.5\textwidth]{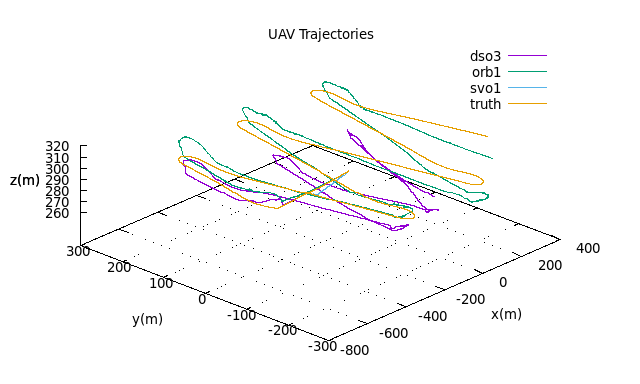}
\caption{\label{fig:trajectory_10mps_25dps_300m_default}Simulation trajectories at 10 m/s, 25 deg/s, 300 m for DSO, SVO, and ORB-SLAM2 compared to truth.}
\end{figure}

\begin{figure*}[t]
\centering
\includegraphics[width=\textwidth]{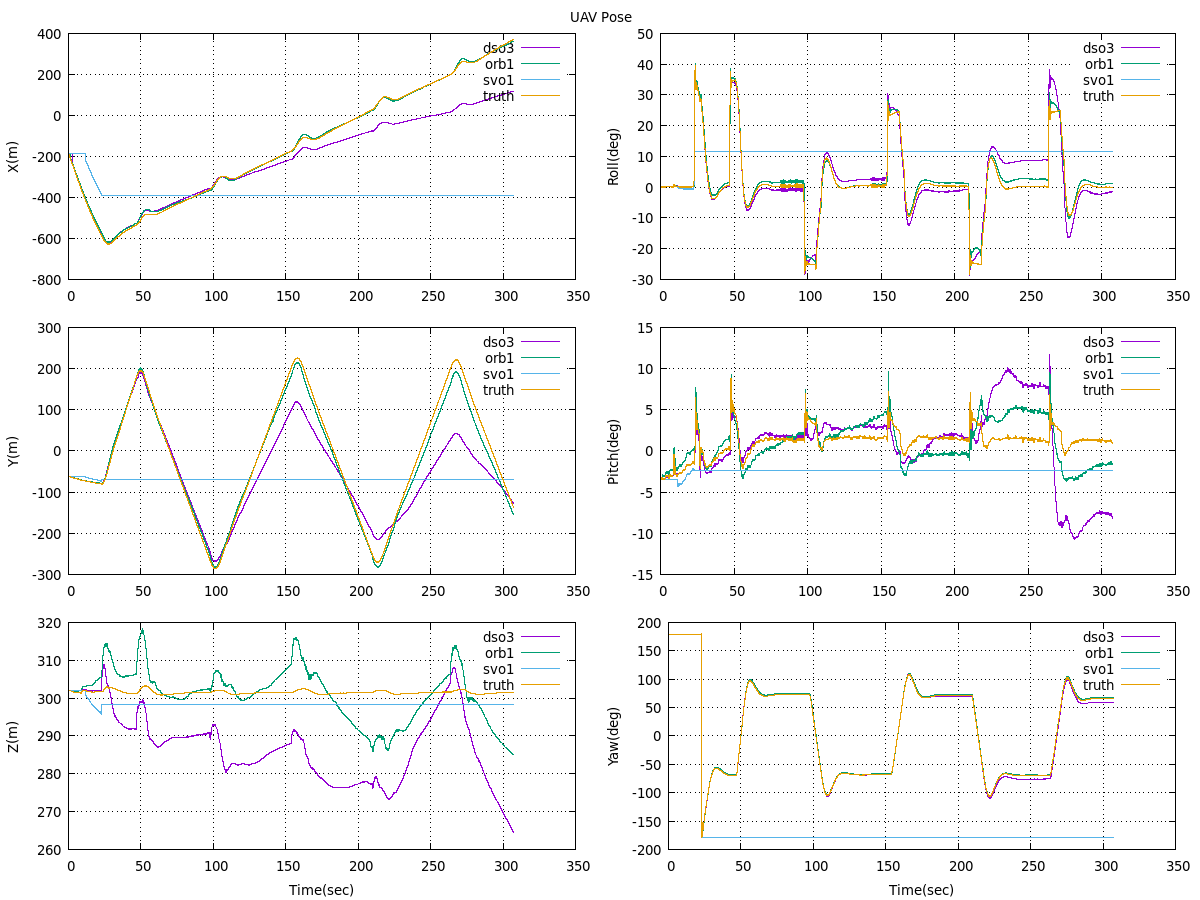}
\caption{\label{fig:pose_10mps_25dps_300m_default}Simulation pose data at 10 m/s, 25 deg/s, 300 m. Note that the scales are different for each plot. For example, the scale of the z position plot is much smaller than that of x or y, greatly emphasizing the difference in values.}
\end{figure*}

Figures \ref{fig:trajectory_10mps_25dps_300m_default} and \ref{fig:pose_10mps_25dps_300m_default} show a representative trajectory from this dataset along with its individual pose data to illustrate the VO performances in detail. This data is obtained from the SUAS flying at 300 m in altitude with a maximum roll rate of 25 deg/s. SVO fails 23 seconds into the trajectory before entering the first turn. DSO drifts further from the truth's position than ORB-SLAM2 as the trajectory progresses, while both DSO and ORB-SLAM2 maintain close approximations of the rotation.

Table \ref{tab:default_25mps} shows the VO performance results at a higher airspeed of 25 m/s. The operational limit at which the algorithms were able to maintain tracking throughout the trajectories was at the maximum roll rate of 45 deg/s for this speed. ORB-SLAM2 was able to maintain localization throughout most of the trajectories at this roll rate while DSO experienced frequent failures and SVO continued to fail on the first turns. All algorithms had difficulty maintaining localization at roll rates of 65 deg/s. ORB-SLAM2 continued to maintain a more accurate position estimate than DSO. DSO exhibited a slightly more accurate rotation estimate at lower roll rates of 25 deg/s although this was reversed at 45 deg/s with ORB-SLAM2 providing significantly better estimates.

\begin{table}[h]
\centering
\caption{\label{tab:default_25mps}Simulation VO performance at 25 m/s. The position and rotation errors for trajectories in which a VO algorithm failed to initialize are marked with an X. The algorithm with the lowest position or rotation error for a trajectory is highlighted.}
\resizebox{0.5\textwidth}{!}{
\begin{tabular}{|c|c|c|c|c|c|c|c|}
\hline
                                                                                       &                                                                                   & \multicolumn{3}{c|}{\textbf{Position RMSE (m)}}                                  & \multicolumn{3}{c|}{\textbf{Rotation RMSE (m)}}                                  \\ \cline{3-8} 
\multirow{-2}{*}{\textbf{\begin{tabular}[c]{@{}c@{}}Roll Rate\\ (deg/s)\end{tabular}}} & \multirow{-2}{*}{\textbf{\begin{tabular}[c]{@{}c@{}}Altitude\\ (m)\end{tabular}}} & \textbf{DSO}                    & \textbf{ORB}                    & \textbf{SVO} & \textbf{DSO}                    & \textbf{ORB}                    & \textbf{SVO} \\ \hline
                                                                                       & \textbf{50}                                                                       & X                               & X                               & X            & X                               & X                               & X            \\ \cline{2-8} 
                                                                                       & \textbf{100}                                                                      & X                               & X                               & X            & X                               & X                               & X            \\ \cline{2-8} 
                                                                                       & \textbf{150}                                                                      & X                               & \cellcolor[HTML]{FCFF2F}81.206  & X            & X                               & \cellcolor[HTML]{FCFF2F}7.047   & X            \\ \cline{2-8} 
                                                                                       & \textbf{200}                                                                      & 167.809                         & \cellcolor[HTML]{FCFF2F}71.906  & X            & \cellcolor[HTML]{FCFF2F}4.535   & 8.326                           & X            \\ \cline{2-8} 
                                                                                       & \textbf{250}                                                                      & 176.991                         & \cellcolor[HTML]{FCFF2F}65.365  & X            & \cellcolor[HTML]{FCFF2F}23.527  & 31.138                          & X            \\ \cline{2-8} 
                                                                                       & \textbf{300}                                                                      & 177.575                         & \cellcolor[HTML]{FCFF2F}62.717  & 368.72       & \cellcolor[HTML]{FCFF2F}7.012   & 9.573                           & 181.637      \\ \cline{2-8} 
                                                                                       & \textbf{350}                                                                      & 167.853                         & \cellcolor[HTML]{FCFF2F}58.907  & 509.857      & \cellcolor[HTML]{FCFF2F}7.111   & 48.707                          & 182.546      \\ \cline{2-8} 
\multirow{-8}{*}{\textbf{25}}                                                          & \textbf{400}                                                                      & 173.911                         & \cellcolor[HTML]{FCFF2F}40.369  & 514.787      & \cellcolor[HTML]{FCFF2F}7.047   & 13.425                          & 182.213      \\ \hline
                                                                                       & \textbf{50}                                                                       & X                               & X                               & X            & X                               & X                               & X            \\ \cline{2-8} 
                                                                                       & \textbf{100}                                                                      & \cellcolor[HTML]{FCFF2F}212.844 & X                               & X            & \cellcolor[HTML]{FCFF2F}82.186  & X                               & X            \\ \cline{2-8} 
                                                                                       & \textbf{150}                                                                      & X                               & \cellcolor[HTML]{FCFF2F}210.371 & X            & X                               & \cellcolor[HTML]{FCFF2F}79.259  & X            \\ \cline{2-8} 
                                                                                       & \textbf{200}                                                                      & X                               & \cellcolor[HTML]{FCFF2F}100.056 & X            & X                               & \cellcolor[HTML]{FCFF2F}49.045  & X            \\ \cline{2-8} 
                                                                                       & \textbf{250}                                                                      & 381.837                         & \cellcolor[HTML]{FCFF2F}90.671  & X            & 83.847                          & \cellcolor[HTML]{FCFF2F}48.052  & X            \\ \cline{2-8} 
                                                                                       & \textbf{300}                                                                      & 407.325                         & \cellcolor[HTML]{FCFF2F}61.127  & 386.2        & 87.066                          & \cellcolor[HTML]{FCFF2F}11.731  & 182          \\ \cline{2-8} 
                                                                                       & \textbf{350}                                                                      & 196.225                         & \cellcolor[HTML]{FCFF2F}47.466  & 505.137      & 9.248                           & \cellcolor[HTML]{FCFF2F}7.815   & 180.623      \\ \cline{2-8} 
\multirow{-8}{*}{\textbf{45}}                                                          & \textbf{400}                                                                      & 313.935                         & \cellcolor[HTML]{FCFF2F}45.072  & 497.773      & 78.373                          & \cellcolor[HTML]{FCFF2F}41.819  & 189.143      \\ \hline
                                                                                       & \textbf{50}                                                                       & X                               & X                               & X            & X                               & X                               & X            \\ \cline{2-8} 
                                                                                       & \textbf{100}                                                                      & \cellcolor[HTML]{FCFF2F}235.459 & X                               & X            & \cellcolor[HTML]{FCFF2F}59.036  & X                               & X            \\ \cline{2-8} 
                                                                                       & \textbf{150}                                                                      & X                               & \cellcolor[HTML]{FCFF2F}419.035 & X            & X                               & \cellcolor[HTML]{FCFF2F}129.506 & X            \\ \cline{2-8} 
                                                                                       & \textbf{200}                                                                      & 444.963                         & \cellcolor[HTML]{FCFF2F}443.574 & X            & \cellcolor[HTML]{FCFF2F}88.568  & 185.526                         & X            \\ \cline{2-8} 
                                                                                       & \textbf{250}                                                                      & \cellcolor[HTML]{FCFF2F}482.253 & 519.21                          & X            & \cellcolor[HTML]{FCFF2F}90.544  & 188.444                         & X            \\ \cline{2-8} 
                                                                                       & \textbf{300}                                                                      & \cellcolor[HTML]{FCFF2F}390.829 & 426.205                         & 506.941      & \cellcolor[HTML]{FCFF2F}86.453  & 118.167                         & 187.349      \\ \cline{2-8} 
                                                                                       & \textbf{350}                                                                      & 513.123                         & \cellcolor[HTML]{FCFF2F}194.008 & 506.12       & 181.218                         & \cellcolor[HTML]{FCFF2F}89.653  & 182.838      \\ \cline{2-8} 
\multirow{-8}{*}{\textbf{65}}                                                          & \textbf{400}                                                                      & \cellcolor[HTML]{FCFF2F}499.621 & 508.358                         & 498.452      & \cellcolor[HTML]{FCFF2F}179.734 & 186.319                         & 180.706      \\ \hline
\end{tabular}}
\end{table}

The simulation experiments clearly identified that ORB-SLAM2 provided the most robust capability to adapting to the new domain of fixed-wing flight.  But the simulation framework also allowed for accomplishing more than identifying the best candidate VO algorithm for use in fixed-wing environments. Hundreds of runs in different configurations and slightly different trajectories were run within days. A comparable study in real life, using live camera data from an aircraft would require a tremendous amount of resources; it would be cost prohibited.  With the virtual world we were also able to hold many conditions constant, like lighting conditions, timing, and the actual trajectory. As will be discussed further below, this is a key feature for the simulation.   

\subsection{Real-World Flight Test}
\begin{figure}[t]
\centering
\includegraphics[width=0.5\textwidth]{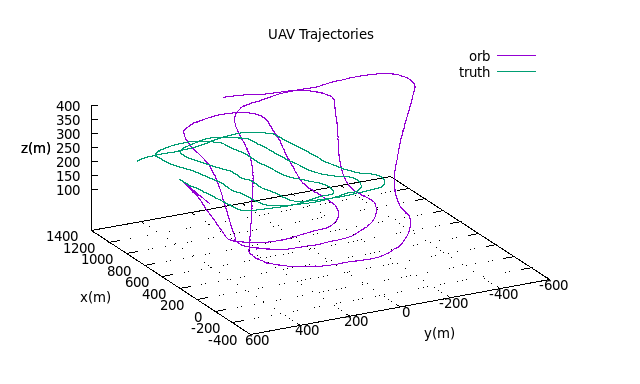}
\caption{\label{fig:real_world_orb_slam_trajectory}Real-world ORB-SLAM2 trajectory compared to truth. }
\end{figure}

\begin{figure*}[t]
\centering
\includegraphics[width=\textwidth]{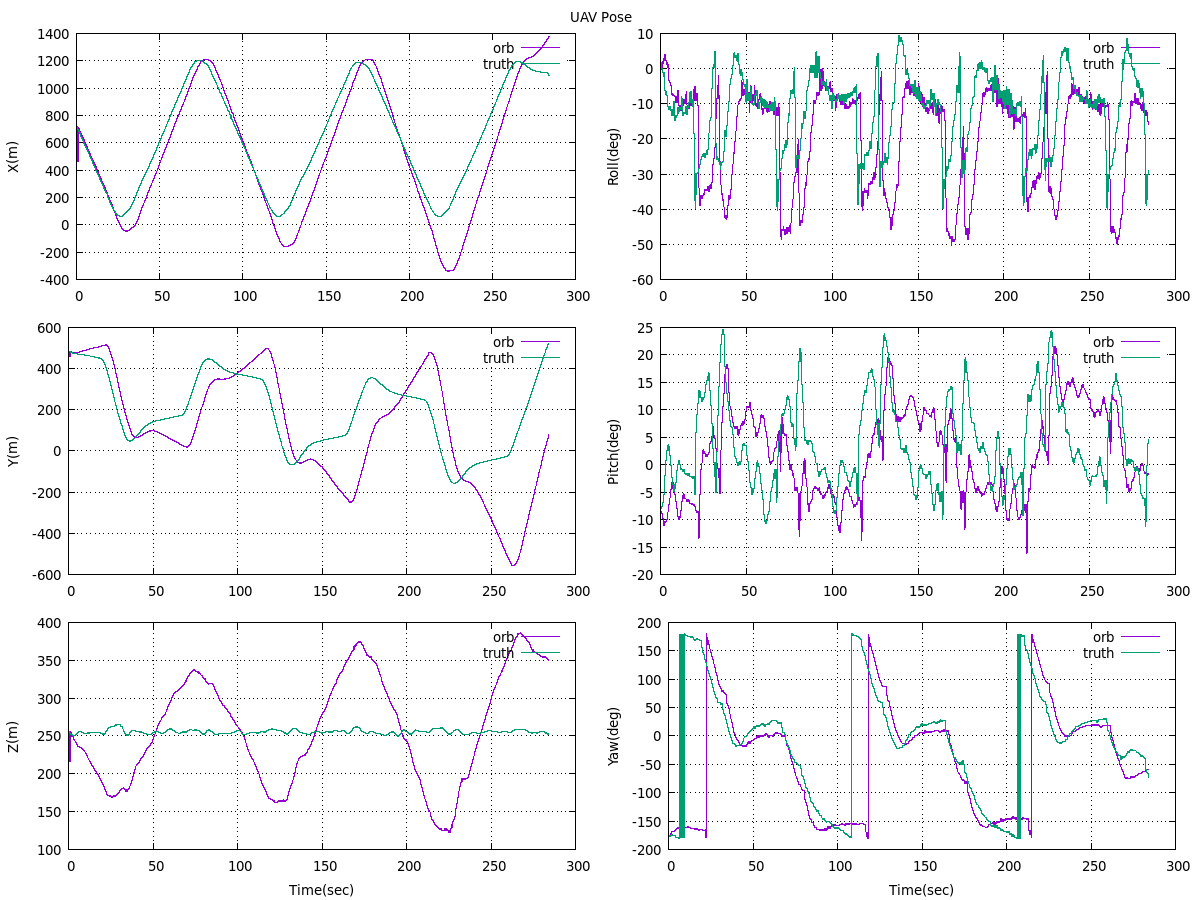}
\caption{\label{fig:real_world_orb_slam_pose}Real-world ORB-SLAM2 pose data results, compared to truth. The most extreme errors occur in the z axis since the algorithm initially miscalculated the pitch, leading to oscillations in altitude between the north and south ends of the trajectory. }
\end{figure*}

All three algorithms experienced difficulties initializing and tracking throughout the gathered real-world dataset due to the previously discussed problem of image saturation. DSO failed to initialize throughout any of the real-world imagery. SVO was able to initialize but failed to maintain tracking for more than a few seconds. ORB-SLAM2 was the most successful and was able to track through most of the trajectory, but still did not manage to get an ideal initialization. Figure \ref{fig:real_world_orb_slam_trajectory} illustrates ORB-SLAM2's estimated trajectory against the truth trajectory data. Figure \ref{fig:real_world_orb_slam_pose} shows the detailed pose estimates. The most extreme errors occur in the z axis since the algorithm initially miscalculated the pitch, leading to oscillations in altitude between the north and south ends of the trajectory. The total position RMSE was 321 m and the rotation RMSE was 112 deg for ORB-SLAM2.  

\begin{figure}[t]
\centering
\includegraphics[width=0.5\textwidth]{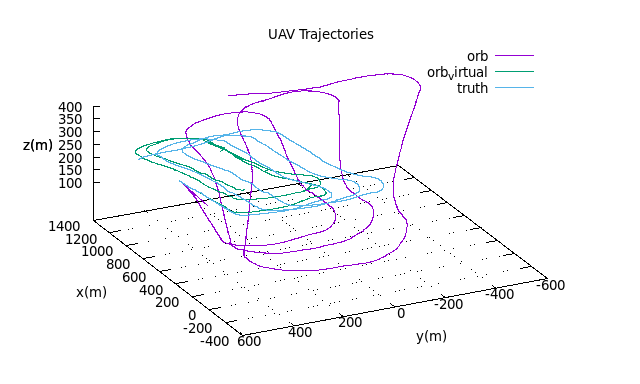}
\caption{\label{fig:real_world_orb_virtual_trajectory}ORB-SLAM2 trajectory from real-world trajectory and Virtual Imagery}
\end{figure}

\begin{figure*}[t]
\centering
\includegraphics[width=\textwidth]{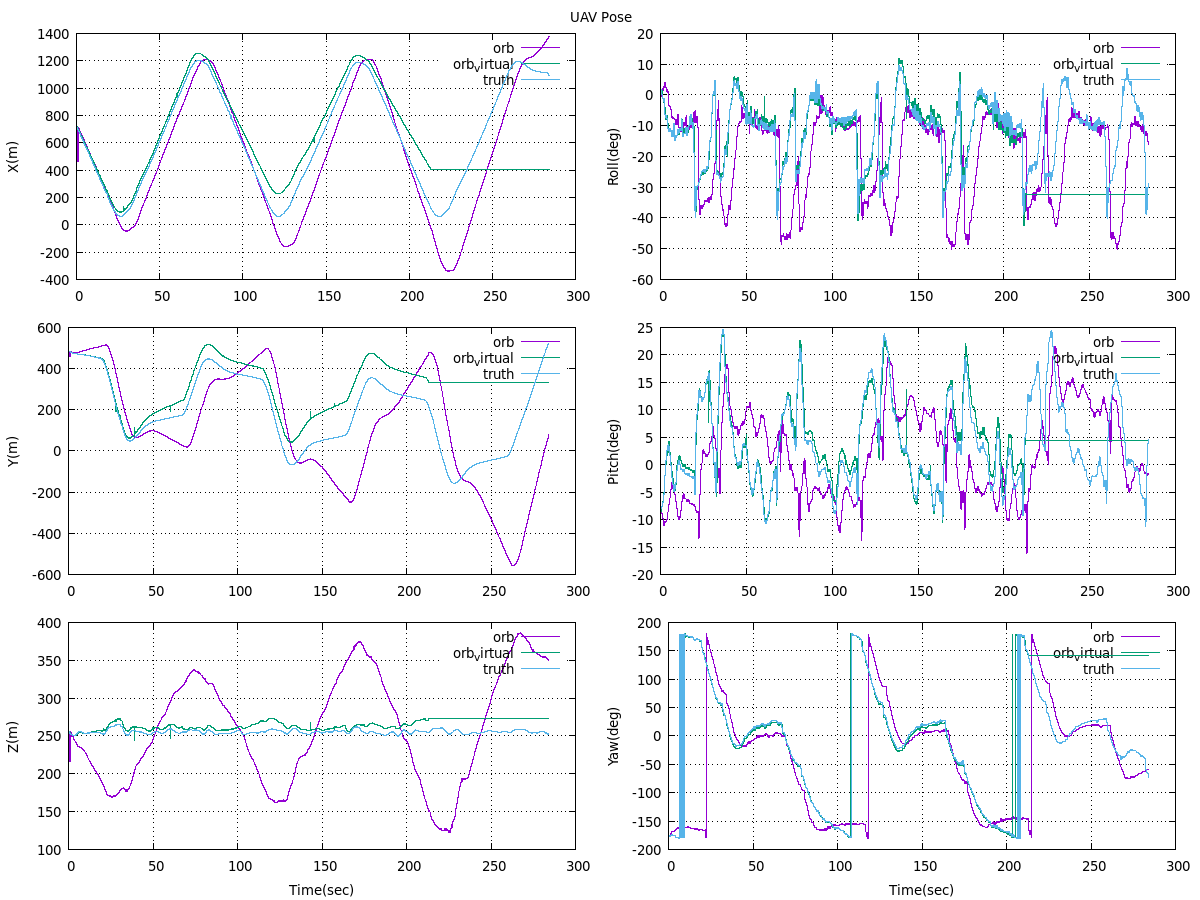}
\caption{\label{fig:real_world_orb_virtual_pose_data}ORB-SLAM2 pose data from real-world trajectory and virtual imagery. The algorithm was able to obtain better initialization using the virtual imagery since the lighting conditions were constant and easier to detect features. However, it lost localization after the second loop due to erratic jumps, likely due to the combinations vibrations and sensor noise,  in the trajectory data}
\end{figure*}

For comparison, ORB-SLAM2 was also run on virtual imagery from the SUAS flying the real-world trajectory dataset in the virtual environment. Both trajectories estimated from the virtual and real-world imagery are shown in Figure \ref{fig:real_world_orb_virtual_trajectory} and their pose data is shown in Figure \ref{fig:real_world_orb_virtual_pose_data}. The algorithm was able to obtain better initialization using the virtual imagery since the lighting conditions were constant and easier to detect features. However, it lost localization after the second loop due to erratic jumps, likely due to the combinations vibrations and sensor noise,  in the trajectory data. To draw better comparisons of algorithm performance between the virtual and real world, ORB-SLAM2 needs to be run on trajectory data that is smoothed and gathered at higher frequencies along with imagery corrected for saturation.

\subsection{Analysis}
Overall, ORB-SLAM2 proves to be the most flexible VO algorithm that is suitable for fixed-wing applications. Both simulation and real-world flight tests show a robust initializer and tracking capability. Although it is not able to achieve the frame rates of SVO or DSO, ORB-SLAM2 is able to track far more robustly at higher speeds and sharper turns, where the combination of 25 m/s and 45 deg/s is the operational limit. Its robustness is further highlighted by its capability to maintain localization under a greatly throttled image stream rate. ORB-SLAM2 also has the distinct advantage in further improvement since a SLAM scheme is incorporated naturally in the design of the algorithm.

SVO is fast but proves to be very fragile, especially during rotations. SVO is unable to cope with any of the turns during simulation or real-world flight tests, even at the slowest speeds and roll rates. This is primarily due to its method of designating new keyframes by a measure of distance. If using SVO for fixed-wing applications, this heuristic must be modified to another method such as that of ORB-SLAM2 or DSO where keyframe creation is based on the measure of changes in its field of view.

DSO's greatest weakness lies in the initializer. The authors note on their git repository\footnote{J. Engel, 'Direct Sparse Odometry', 2018. [Online]. Available: https://github.com/JakobEngel/dso. [Accessed: 23- Jan- 2019].} that the current initializer is not very good and requires slow, easy movements with lots of translation and little rotation. This is much easier to achieve in simulation with perfectly smooth flights creating ideal environments for initialization. However, real-world flights prove to be much more difficult as the SUAS invariably encounters turbulence from wind. This is shown by DSO being unable to successfully initialize in any of the real-world datasets. If using DSO, the initializer should be replaced by a faster method such as through feature-based homography or RANSAC. The initializer could also be either augmented or completely replaced by IMU measurements. Another drawback to DSO is that it is not as amenable to incorporating loop closures as other feature-based VO algorithms, since many of the current loop closure frameworks make use of features for the bag-of-words approach \cite{Cummins2008}.


\section{Conclusion}
In this paper, we presented a virtual environment for real-time testing and analysis of VO and VSLAM algorithms. This was used to test the performance of three open-source algorithms: DSO, SVO, and ORB-SLAM2 (with loop closures disabled). The algorithms were tested under various airspeeds, roll rates, and altitudes. The algorithms were also tested on a real-world dataset and the results compared against virtual data. Through these tests, ORB-SLAM2 was found to be the most flexible and robust algorithm in both simulation and real-world flights. Although this work demonstrated the flexibility of the virtual testbed, further work must be accomplished to draw better correlations and conclusions between the state estimates obtained from flight tests in the virtual environment and the real world, primarily obtaining a filtered real-world dataset with a more accurate and smoothed ground truth along with imagery corrected for saturation.

\bibliographystyle{SageV}
\bibliography{references}

\begin{thebibliography}{10}
\providecommand{\url}[1]{\texttt{#1}}
\providecommand{\urlprefix}{URL }
\expandafter\ifx\csname urlstyle\endcsname\relax
  \providecommand{\doi}[1]{DOI:\discretionary{}{}{}#1}\else
  \providecommand{\doi}{DOI:\discretionary{}{}{}\begingroup
  \urlstyle{rm}\Url}\fi
\providecommand{\eprint}[2][]{\url{#2}}

\bibitem{Engel2018}
Engel J, Koltun V and Cremers D.
\newblock {Direct Sparse Odometry}.
\newblock \emph{IEEE Transactions on Pattern Analysis and Machine Intelligence}
  2018; 40(3): 611--625.
\newblock \doi{10.1109/TPAMI.2017.2658577}.

\bibitem{Forster2014svo}
Forster C, Pizzoli M and Scaramuzza D.
\newblock {SVO: Fast semi-direct monocular visual odometry}.
\newblock \emph{Proceedings - IEEE International Conference on Robotics and
  Automation} 2014; (May): 15--22.
\newblock \doi{10.1109/ICRA.2014.6906584}.

\bibitem{Mur-Artal2017}
Mur-Artal R and Tardos JD.
\newblock {ORB-SLAM2: An Open-Source SLAM System for Monocular, Stereo, and
  RGB-D Cameras}.
\newblock \emph{IEEE Transactions on Robotics} 2017; 33(5): 1255--1262.
\newblock \doi{10.1109/TRO.2017.2705103}.

\bibitem{Hartley2004}
Hartley R and Zisserman R.
\newblock \emph{{Multiple View Geometry in Computer Vision}}.
\newblock 2 ed. Cambridge U.K: Cambridge Univ. Press, 2004.
\newblock ISBN 9780521540513.

\bibitem{Scaramuzza2011b}
Scaramuzza D and Fraundorfer F.
\newblock {Visual Odometry}.
\newblock \emph{IEEE Robotics {\&} Automation Magazine} 2011; (December).

\bibitem{Scaramuzza2011}
Scaramuzza D and Fraundorfer F.
\newblock {Visual Odometry Part II}.
\newblock \emph{IEEE Robotics {\&} Automation Magazine} 2011; 18(4): 80--92.
\newblock \doi{10.1109/MRA.2011.943233}.

\bibitem{Durrant-whyte2006}
Durrant-Whyte H and Bailey T.
\newblock {Simultaneous localization and mapping ({\{}SLAM{\}}): part
  {\{}II{\}}}.
\newblock \emph{Robotics and Automation Magazine} 2006; 13(3): 99--110.
\newblock \doi{10.1109/MRA.2006.1678144}.

\bibitem{bailey06simultaneous}
Bailey T and Durrant-Whyte H.
\newblock {Simultaneous localization and mapping ({\{}SLAM{\}}): part
  {\{}II{\}}}.
\newblock \emph{IEEE Robotics {\&} Automation Magazine} 2006; 13(3): 108--117.
\newblock \doi{10.1109/MRA.2006.1678144}.

\bibitem{Hassaballah2016}
Hassaballah M, Abdelmgeid AA and Alshazly HA.
\newblock \emph{{Image Feature Detectors and Descriptors}}, volume 630.
\newblock 2016.
\newblock ISBN 978-3-319-28852-9.
\newblock \doi{10.1007/978-3-319-28854-3}.

\bibitem{Irani1999}
Irani M and Anandan P.
\newblock {All About Direct Methods}.
\newblock \emph{ICCV workshop on Vision Algorithms} 1999; : 267--277.

\bibitem{Klein2007}
Klein G and Murray D.
\newblock {Parallel tracking and mapping for small AR workspaces}.
\newblock In \emph{Proceedings of the 2007 6th IEEE and ACM International
  Symposium on Mixed and Augmented Reality}.

\bibitem{Newcombe2011}
Newcombe RA, Lovegrove SJ and Davison AJ.
\newblock {DTAM : Dense Tracking and Mapping in Real-Time}.
\newblock \emph{Proceedings of the International Conference on Computer Vision
  (ICCV)} 2011; .

\bibitem{Engel2014lsd}
Engel J, Sch T and Cremers D.
\newblock {LSD-SLAM: Large-scale Direct Monocular SLAM}.
\newblock In \emph{European Conference on Computer Vision}. pp. 834--849.

\bibitem{Engel2016}
Engel J, Usenko V and Cremers D.
\newblock {A Photometrically Calibrated Benchmark For Monocular Visual
  Odometry}.
\newblock \emph{Computing Research Repository (CoRR)} 2016;
  \doi{10.1016/j.aqpro.2013.07.003}.

\bibitem{Mur-Artal2015}
Mur-Artal R, Montiel JM and Tardos JD.
\newblock {ORB-SLAM: A Versatile and Accurate Monocular SLAM System}.
\newblock \emph{IEEE Transactions on Robotics} 2015; 31(5): 1147--1163.
\newblock \doi{10.1109/TRO.2015.2463671}.

\bibitem{Pizzoli2014}
Pizzoli M, Forster C and Scaramuzza D.
\newblock {REMODE: Probabilistic, monocular dense reconstruction in real time}.
\newblock \emph{Proceedings - IEEE International Conference on Robotics and
  Automation} 2014; : 2609--2616\doi{10.1109/ICRA.2014.6907233}.

\bibitem{Faessler2015}
Faessler M, Fontana F, Forster C et~al.
\newblock {Autonomous, Vision-based Flight and Live Dense 3D Mapping with a
  Quadrotor Micro Aerial Vehicle}.
\newblock \emph{J Field Robotics} 2015; 23(0): 1–20.
\newblock \doi{10.1002/rob}.

\bibitem{Forster2017}
Forster C, Zhang Z, Gassner M et~al.
\newblock {Semi-Direct Visual Odometry for Monocular , Wide-angle, and
  Muti-Camera Systems}.
\newblock \emph{IEEE Transactions on Robotics} 2017; 33: 249--265.
\newblock \doi{10.1109/TRO.2016.2623335}.

\bibitem{Wheeler2017}
Wheeler DO and Koch DP.
\newblock {Relative Navigation of Autonomous GPS- Degraded Micro Air Vehicles}.
\newblock \emph{IEEE Control Systems Magazine} 2018; 38(4): 30--48.

\bibitem{Ellingson2018}
Ellingson G, Brink K and McLain T.
\newblock {Relative visual-inertial odometry for fixed-wing aircraft in
  GPS-denied environments}.
\newblock \emph{2018 IEEE/ION Position, Location and Navigation Symposium
  (PLANS)} 2018; : 786--792\doi{10.1109/PLANS.2018.8373454}.
\newblock \urlprefix\url{https://ieeexplore.ieee.org/document/8373454/}.

\bibitem{Wang2017DeepVO:Networks}
Wang S, Clark R, Wen H et~al.
\newblock {DeepVO: Towards end-to-end visual odometry with deep Recurrent
  Convolutional Neural Networks}.
\newblock \emph{Proceedings - IEEE International Conference on Robotics and
  Automation} 2017; : 2043--2050\doi{10.1109/ICRA.2017.7989236}.

\bibitem{Geiger2013}
Geiger A, Lenz P, Stiller C et~al.
\newblock {Vision meets Robotics: The KITTI Dataset}.
\newblock \emph{International Journal of Robotics Research (IJRR)} 2013;
  (October): 1--6.

\bibitem{Sturm2012}
Sturm J, Engelhard N, Endres F et~al.
\newblock {A Benchmark for the Evaluation of RGB-D SLAM Systems}.
\newblock \emph{Inernational Conference on Intelligent Robot Systems (IROS)}
  2012; .

\bibitem{Burri2016a}
Burri M, Nikolic J, Gohl P et~al.
\newblock {The EuRoC micro aerial vehicle datasets}.
\newblock \emph{International Journal of Robotics Research} 2016; 35(10):
  1157--1163.
\newblock \doi{10.1177/0278364915620033}.

\bibitem{Majdik2016}
Majdik A, Till C and Scaramuzza D.
\newblock {The Zurich Urban Micro Aerial Vehicle Dataset}.
\newblock \emph{I J Robotics Res} 2017; 36: 269--273.

\bibitem{gaidon2016virual}
Gaidon A, Wang Q, Cabon Y et~al.
\newblock {VirtualWorlds as Proxy for Multi-object Tracking Analysis}.
\newblock \emph{Proceedings of the IEEE Computer Society Conference on Computer
  Vision and Pattern Recognition} 2016; 2016-Decem: 4340--4349.
\newblock \doi{10.1109/CVPR.2016.470}.

\bibitem{sayre2018visual}
Sayre-Mccord T, Guerra W, Antonini A et~al.
\newblock {Visual-Inertial Navigation Algorithm Development Using
  Photorealistic Camera Simulation in the Loop}.
\newblock \emph{Proceedings - IEEE International Conference on Robotics and
  Automation} 2018; : 2566--2573\doi{10.1109/ICRA.2018.8460692}.

\bibitem{shah2018airsim}
Shah S, Dey D, Lovett C et~al.
\newblock {AirSim: High-Fidelity Visual and Physical Simulation for Autonomous
  Vehicles}.
\newblock In Hutter M and Siegwart R (eds.) \emph{Field and Service Robotics}.
  Cham: Springer International Publishing.
\newblock ISBN 978-3-319-67361-5, pp. 621--635.

\bibitem{zhang2016benefit}
{Zichao Zhang}, Rebecq H, Forster C et~al.
\newblock {Benefit of large field-of-view cameras for visual odometry}.
\newblock In \emph{2016 IEEE International Conference on Robotics and
  Automation (ICRA)}. pp. 801--808.
\newblock \doi{10.1109/ICRA.2016.7487210}.

\bibitem{koenig2004design}
Koenig N and Howard A.
\newblock {Design and use paradigms for Gazebo, an open-source multi-robot
  simulator}.
\newblock In \emph{2004 IEEE/RSJ International Conference on Intelligent Robots
  and Systems (IROS) (IEEE Cat. No.04CH37566)}, volume~3. pp. 2149--2154.
\newblock \doi{10.1109/IROS.2004.1389727}.

\bibitem{furrer2016rotors}
Furrer F, Burri M, Achtelik MW et~al.
\newblock \emph{{RotorS – A Modular Gazebo MAV Simulator Framework}}, volume
  625.
\newblock 2016.
\newblock ISBN 978-3-319-26052-5.
\newblock \doi{10.1007/978-3-319-26054-9}.
\newblock
  \urlprefix\url{http://www.scopus.com/inward/record.url?eid=2-s2.0-84958580682&partnerID=tZOtx3y1}.

\bibitem{Nykl2008}
Nykl S, Mourning C, Leitch M et~al.
\newblock {An Overview of the STEAMiE Educational Game Engine}.
\newblock \emph{Proceedings - Frontiers in Education Conference, FIE} 2008; :
  21--25\doi{10.1109/FIE.2008.4720454}.

\bibitem{Johnson2017}
Johnson DT, Nykl SL and Raquet JF.
\newblock {Combining Stereo Vision and Inertial Navigation for Automated Aerial
  Refueling}.
\newblock \emph{Journal of Guidance, Control, and Dynamics} 2017; 40(9):
  2250--2259.
\newblock \doi{10.2514/1.g002648}.

\bibitem{nykl2018_boomOcclusion}
Paulson Z, Nykl S, John P et~al.
\newblock {Mitigating the Effects of Boom Occlusion on Automated Aerial
  Refueling through Shadow Volumes}.
\newblock \emph{The Journal of Defense Modeling and Simulation} 2018; 0(0):
  1--15.
\newblock \doi{10.1177/1548512918808408}.

\bibitem{roeber2018_SfM}
Roeber J, Nykl S and Graham~F S.
\newblock {Assessment of Structure from Motion for Reconnaissance Augmentation
  and Bandwidth Usage Reduction}.
\newblock \emph{Journal of Defense Modeling and Simulation} 2019; 1(1): 1--13.
\newblock \doi{10.1177/1548512919844021}.

\bibitem{parsonsAAR_AIAA_IS}
Parsons C, Paulson Z, Nykl SL et~al.
\newblock {Analysis of Simulated Imagery for Real-Time Vision-Based Automated
  Aerial Refueling}.
\newblock \emph{AIAA: Journal of Aerospace Information Systems} 2019; :
  1--14\doi{10.2514/1.I010634}.

\bibitem{Quigley2009}
Quigley M, Conley K, Gerkey B et~al.
\newblock {ROS: an open-source Robot Operating System}.
\newblock \emph{Icra} 2009; 3(Figure 1): 5.
\newblock
  \doi{http://www.willowgarage.com/papers/ros-open-source-robot-operating-system}.

\bibitem{Huang10LCM}
Huang AS, Olson E and Moore DC.
\newblock {LCM: Lightweight Communications and Marshalling}.
\newblock In \emph{IEEE/RSJ 2010 International Conference on Intelligent Robots
  and Systems, IROS 2010}.
\newblock ISBN 9781424466757, pp. 4057--4062.
\newblock \doi{10.1109/IROS.2010.5649358}.

\bibitem{Kim2019Monocular}
Kim KM.
\newblock \emph{{Monocular Visual Odometry for Fixed-Wing Small Unmanned
  Aircraft Systems}}.
\newblock PhD Thesis, Air Force Institute of Technology, 2019.

\bibitem{Cummins2008}
Cummins M and Newman P.
\newblock {{\{}FAB-MAP{\}}: Probabilistic Localization and Mapping in the Space
  of Appearance}.
\newblock \emph{The International Journal of Robotics Research} 2008; 27(6):
  647--665.
\newblock \doi{10.1177/0278364908090961}.
\newblock
  \urlprefix\url{http://ijr.sagepub.com/cgi/doi/10.1177/0278364908090961
  http://dx.doi.org/10.1177/0278364908090961}.

\end{thebibliography}

\end{document}